\documentclass[journal]{IEEEtran}

\ifCLASSINFOpdf
\else
\fi

\newenvironment{sloppypar*}
{\sloppy\ignorespaces}
{\par}
\usepackage[T1]{fontenc}
\usepackage{balance}
\usepackage{graphicx}
\usepackage{amsmath}
\usepackage{wrapfig}
\usepackage{adjustbox}  
\usepackage{hyperref}
\usepackage{subfig}
\usepackage{cmap}
\usepackage[ruled,linesnumbered]{algorithm2e}
\SetKwInput{KwInput}{Input}                
\SetKwInput{KwOutput}{Output}              

\ifodd 1

\else

\fi

\newcommand{\name}{GAPSL\xspace}



\usepackage{physics}
\usepackage{mathtools}
\usepackage{cmap}
\usepackage{multirow}
\usepackage{placeins}
\usepackage[utf8]{inputenc} 
\usepackage{hyperref}       
\usepackage{url}            
\usepackage{booktabs}       
\usepackage{amsfonts}       
\usepackage{nicefrac}       
\usepackage{microtype}      
\usepackage{xcolor}         
\usepackage{booktabs}
\usepackage[table]{xcolor}  

\begin{document}

\title{\name: A Gradient-Aligned Parallel Split Learning over Data-Heterogeneous Edge Computing Systems}

\author{Zheng Lin,~\IEEEmembership{Member,~IEEE},   
Ons Aouedi,~\IEEEmembership{Member,~IEEE}, Zihan Fang, Wei Ni,~\IEEEmembership{Fellow,~IEEE}, Yue Gao,~\IEEEmembership{Fellow,~IEEE}, Symeon Chatzinotas,~\IEEEmembership{Fellow,~IEEE}, and Xianhao Chen,~\IEEEmembership{Member,~IEEE}

\thanks{Z. Lin and X. Chen are with the Department of Electrical and Computer Engineering, University of Hong Kong, Pok Fu Lam, Hong Kong, China (e-mail: zhenglin@ieee.org; xchen@eee.hku.hk).}
\thanks{Z. Fang is with Hong Kong JC STEM Lab of Smart City and Department of Computer Science, City University of Hong Kong, Kowloon, Hong Kong SAR, China (e-mail: zihanfang3-c@my.cityu.edu.hk).}
\thanks{W. Ni is with the School of Engineering, Edith Cowan University, Perth, WA 6027, Australia (email: wei.ni@ieee.org).}
\thanks{Y. Gao is with the Institute of Space Internet, Fudan University, Shanghai 200438, China (email: gao.yue@fudan.edu.cn).} 
\thanks{O. Aouedi and S. Chatzinotas are with the Interdisciplinary Centre for Security, Reliability and Trust (SnT), University of Luxembourg, L-1855 Luxembourg, Luxembourg (e-mail: ons.aouedi@uni.lu; symeon.chatzinotas@uni.lu).}
}


\maketitle

{\begin{abstract}
The increasing complexity of neural networks poses significant challenges for democratizing federated learning (FL) on resource-constrained edge devices. Parallel split learning (PSL) has emerged as a promising solution by offloading substantial computing workload to a server via model partitioning, shrinking client-side computing load, and eliminating the client-side model aggregation for reduced communication and deployment costs. However, the highly heterogeneous nature of client data in edge computing systems causes \textit{aggregation-free} PSL to suffer from severe training divergence, stemming from gradient directional inconsistency across clients. To address this challenge, we propose \name, a gradient-aligned PSL framework tailored for data-heterogeneous edge systems, which comprises two key components: leader gradient identification (LGI) and gradient direction alignment (GDA). LGI dynamically selects a set of directionally consistent device gradients to construct a leader gradient as a robust proxy for the global convergence trend.  GDA employs a direction-aware regularization to align each client’s gradient with the leader gradient, thereby mitigating inter-device gradient directional inconsistency and enhancing model convergence. We evaluate \name on a prototype computing testbed. Extensive experiments demonstrate that \name consistently outperforms state-of-the-art benchmarks in training accuracy, convergence latency, and system robustness under severe data heterogeneity.

\end{abstract}}


\IEEEpeerreviewmaketitle

\section{Introduction}\label{sec:introduction} 
The Internet of Things (IoT) is experiencing rapid expansion, interconnecting billions of smart devices that continuously generate vast volumes of privacy-sensitive data~\cite{liu2025fedmobile,fang2025dynamic,lin2025hasfl,sarhaddi2025llms}. This unprecedented amount of data significantly advances the maturity of machine learning (ML) for enabling intelligent IoT applications. Conventional ML predominantly relies on centralized learning (CL), where raw data is collected and processed on a central server for model training~\cite{li2025unleashing,subramanya2021centralized,lin2024fedsn,ottun2022social,fang2024automated}. With the exponential growth of global data traffic—projected to reach 181 zettabytes by 2025 and 394 zettabytes by 2028~\cite{Statista}—CL faces serious limitations in communication overhead and data privacy risks. Transmitting such large volumes of raw data over dynamic wireless networks incurs prohibitive communication overhead, and centralized storage and processing of sensitive data raise even more critical privacy concerns~\cite{liu2023optimizing,deng2022actions}. To address these issues, federated learning (FL)~\cite{mcmahan2017communication,solans2024non} has emerged as an alternative that enables collaborative model training across edge devices by exchanging model parameters rather than raw data~\cite{chen2024optimizing,zheng2024trust}. While FL alleviates privacy and bandwidth concerns, it encounters scalability bottlenecks as ML models scale up~\cite{lin2025hsplitlora,wu2023split}. For instance, training and deploying large-sized models, such as Gemini Nano-2 with 3.25 billion parameters (3GB for 32-bit floats), can exceed the computing and storage capabilities of resource-constrained edge devices~\cite{team2023gemini}.

\begin{figure}[t]
\centering
\includegraphics[width=0.90\columnwidth]{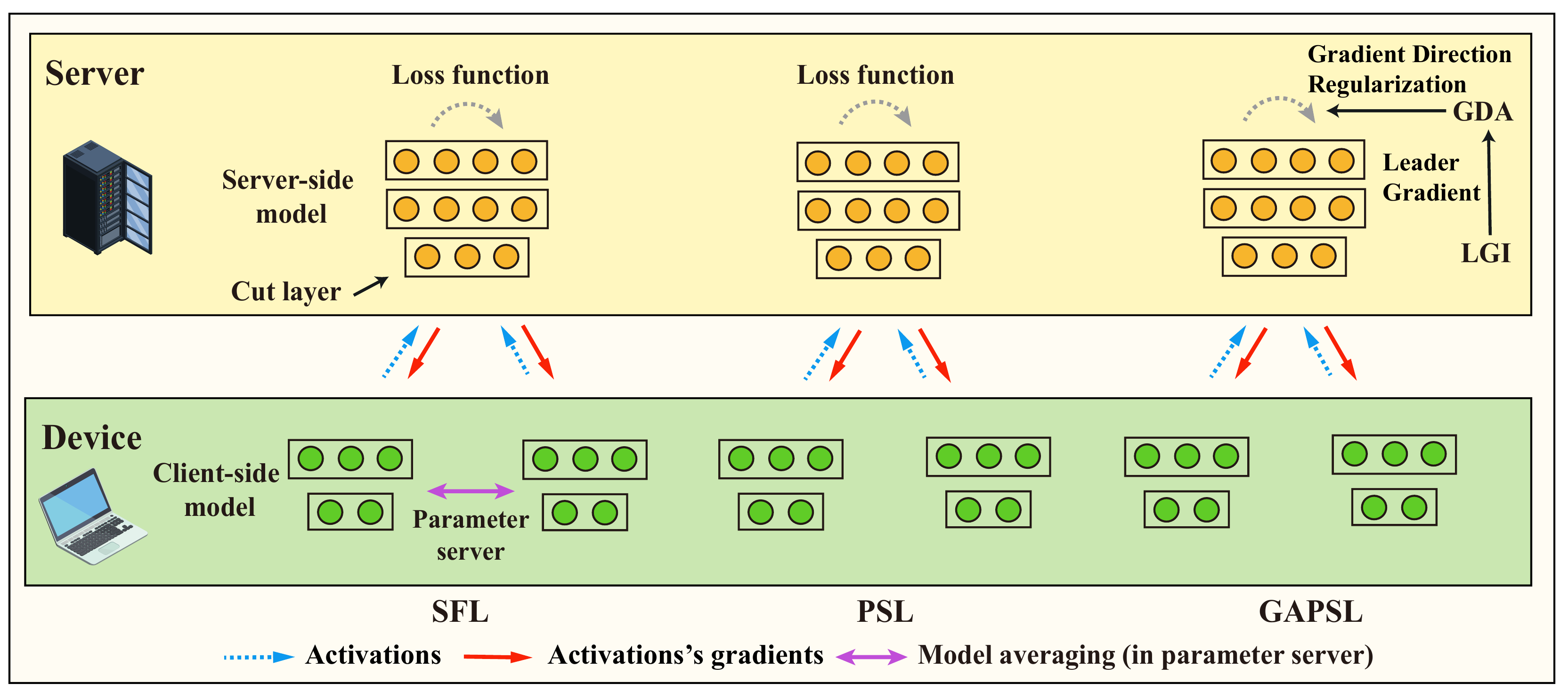}
\vspace{-0.1em}
\caption{
The comparison of SFL, PSL, and \name frameworks,
where SFL relies on a parameter server to aggregate client-side models and PSL streamlines the training process by eliminating the client-side model aggregation.
}
\label{fig:compare_variant_SL}
\end{figure}

Split federated learning (SFL)~\cite{thapa2022splitfed} offers a promising solution to addressing the weaknesses of FL by offloading the computation-intensive workload from edge devices to an edge server via model partitioning~\cite{tirana2026data}. By integrating the model partitioning of split learning (SL) with the parallel training paradigm of FL, SFL enables multiple edge devices to collaboratively train a shared model. Despite its advantages, deploying SFL in edge computing systems still faces a severe limitation. It relies on a third-party parameter server\footnote{The parameter server cannot exchange information with the main server. Otherwise, this collusion can lead to severe privacy risks because the server knows both activations and model parameters from clients~\cite{pasquini2021unleashing,lin2023efficient}.} to aggregate client-side models. The periodic model aggregation requires exchanging model parameters over dynamic wireless links, which incurs severe communication latency and imposes strict synchronization barriers that exacerbate straggler effects. Furthermore, from a network management perspective, deploying and maintaining an additional parameter server introduces substantial infrastructural overhead and deployment complexity. To address this issue, parallel split learning (PSL)~\cite{kim2022bargaining,joshi2021splitfed} has been proposed as an aggregation-free variant of SFL. PSL streamlines the training procedure of SFL by eliminating the need for client-side model aggregation, substantially reducing communication overhead and simplifying edge system deployment. A comparative illustration of SFL and PSL is presented in Fig.~\ref{fig:compare_variant_SL}.

While PSL offers a lightweight and communication-efficient alternative to SFL by eliminating client-side model aggregation, it is inherently more vulnerable to data heterogeneity in edge computing systems. In contrast to FL and SFL, where periodic model aggregation serves as a statistical calibration mechanism to mitigate the impact of non-independent and identically distributed (non-IID) data, PSL operates without any such aggregation~\cite{lin2023split}. As a result, each edge device independently generates activations based solely on its local data, which are often severely skewed due to class imbalance or covariate shift~\cite{yu2025towards,li2024fedcir}. These semantically inconsistent and statistically biased activations lead to divergent gradient directions during the back-propagation (BP) process on the server. The resulting gradient updates exhibit a dual nature: while some gradients align with the global convergence direction, others, dominated by device-specific inductive biases, introduce severe noise or even adversarial updates. Consequently, PSL exhibits much higher gradient divergence than SFL due to the lack of client-side model aggregation. Such conflicting gradients without calibration not only fail to correct activation biases but also exacerbate gradient directional inconsistency, ultimately causing poor learning performance or even convergence failure~\cite{deng2024fedasa}. 
We have conducted empirical studies in {{Section~\ref{sec:background_motivation}}} and identified gradient directional inconsistency as an underlying cause of the convergence failure of PSL. This analysis provides key insights and motivation for the design of \name, which explicitly addresses gradient inconsistency to enhance training stability and convergence efficiency.

To tackle the above challenge, as shown in Fig.~\ref{fig:compare_variant_SL}, we propose a
\underline{g}radient-\underline{a}ligned \underline{PSL} framework, named \name, with a gradient-aligned mechanism to mitigate the training divergence caused by conflicting gradients across edge devices. \name comprises two key components. First, to identify a gradient direction reflecting the global convergence trend, we design a leader gradient identification (LGI) mechanism that dynamically selects a set of directionally consistent device gradients to construct a leader gradient. Second, to enhance inter-device gradient alignment and mitigate directional conflicts during training, we develop a gradient direction alignment (GDA) scheme that employs direction-aware regularization to steer each device's gradient toward the leader gradient. The key contributions of this paper are summarized as follows:
\vspace{-0.2em}
\begin{itemize}
    \item We propose \name, a gradient-aligned PSL framework that resolves the problem of gradient directional inconsistency, enabling efficient and stable SL without client-side model aggregation.
    \item We design LGI to dynamically select a set of gradients based on directional consistency and construct a leader gradient as a robust proxy for the global convergence direction.
    \item We devise GDA to adaptively align each client’s gradient with the leader gradient via direction-aware regularization, thereby mitigating gradient directional conflicts and improving model convergence. 
    \item We evaluate \name on a prototype computing testbed with extensive experiments. The results demonstrate that \name outperforms state-of-the-art frameworks in training accuracy, convergence latency, and system robustness against data heterogeneity.
\end{itemize}
\vspace{-0.2em}

The rest of this paper is organized as follows. Section~\ref{sec:background_motivation} motivates the design of \name and Section~\ref{sec:s3l_frame} elaborates on the system design.
Section~\ref{sec:implementation} introduces the system implementation, followed by performance evaluation in Section~\ref{sec:simulation results}.
Related works and technical limitations are discussed in Section~\ref{sec:related_work}.
Finally, conclusions are presented in Section~\ref{sec:conclusion}.

\section{Challenge and Motivation}  \label{sec:background_motivation}

To elucidate the fundamental limitations of PSL and motivate the design of \name, we conduct empirical studies to investigate the impacts of i) the lack of client-side model aggregation and ii) gradient directional inconsistency on training performance.



\subsection{Lack of Client-side Model Aggregation}\label{challenge_1}

While PSL reduces communication and deployment costs by eliminating client-side model aggregation, it inherently lacks model synchronization~\cite{kim2022bargaining,joshi2021splitfed}. Thus, each edge device generates semantically inconsistent and statistically biased activations based solely on its non-IID local dataset, which are directly used to train the model. Without aggregation to calibrate these biases, the semantic discrepancies accumulate as training progresses, destabilizing the learning process and impeding convergence.

\begin{figure}[t]
  \centering
  \vspace{-0.8em}
  \subfloat[Training performance.\label{subfig:motivate_1_1}]{
    \includegraphics[width=0.235\textwidth]{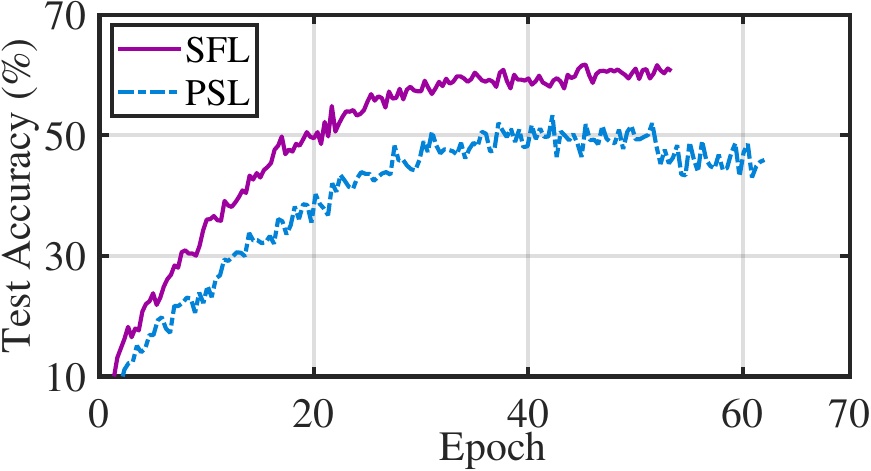}
  }
  \subfloat[Test accuracy.\label{subfig:motivate_1_2}]
  {
    \includegraphics[width=0.236\textwidth]{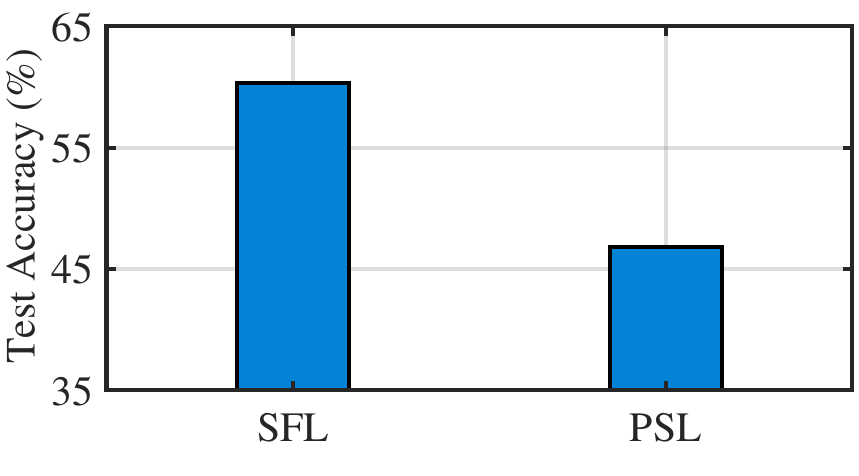}
  }
  \vspace{-0.1em}
  \caption{The training performance (a) and test accuracy (b) of SFL and PSL on CIFAR-10 using VGG-16. }
  \label{fig:motivating_1_total}
\end{figure}

To gain a deeper understanding of the impact of lacking client-side model aggregation in PSL, we compare its performance with SFL using the VGG-16 neural network~\cite{simonyan2014very} on the CIFAR-10 dataset~\cite{krizhevsky2009learning}. Unless otherwise specified, this configuration remains consistent throughout the whole Section~\ref{sec:background_motivation}. Following prior studies~\cite{hu2025faster,lin2025leo,zhang2024multi}, we adopt Dirichlet data partitioning to set the degree of data heterogeneity controlled by parameter $\alpha$, where a smaller $\alpha$ represents a higher degree of data heterogeneity, indicating that each edge device holds data from a more skewed or narrower subset of classes. In our experiment, we set the Dirichlet partitioning parameter $\alpha$ to 0.1. Fig.~\ref{subfig:motivate_1_1} illustrates that PSL exhibits significantly slower and less stable convergence compared to SFL. Crucially, PSL fails to achieve model convergence entirely, indicating a substantial collapse in learning stability under high data heterogeneity, rendering it impractical for edge deployments. Furthermore, Fig.~\ref{subfig:motivate_1_2} reveals that PSL consistently achieves lower test accuracy than SFL. These observations suggest that PSL suffers not only from significantly degraded performance but also from impaired convergence stability in the non-IID setting. In contrast, SFL demonstrates greater robustness due to its periodic client-side model aggregation, which helps mitigate semantic drift across edge devices. Motivated by this, it is imperative to develop coordination mechanisms for PSL that enforce semantic consistency across clients, thereby improving both convergence robustness and model performance.

\begin{figure}[t]
\vspace{-0.8em}
  \centering
  \subfloat[Test accuracy.\label{subfig:motivate_2_1}]{
    \includegraphics[width=0.233\textwidth]{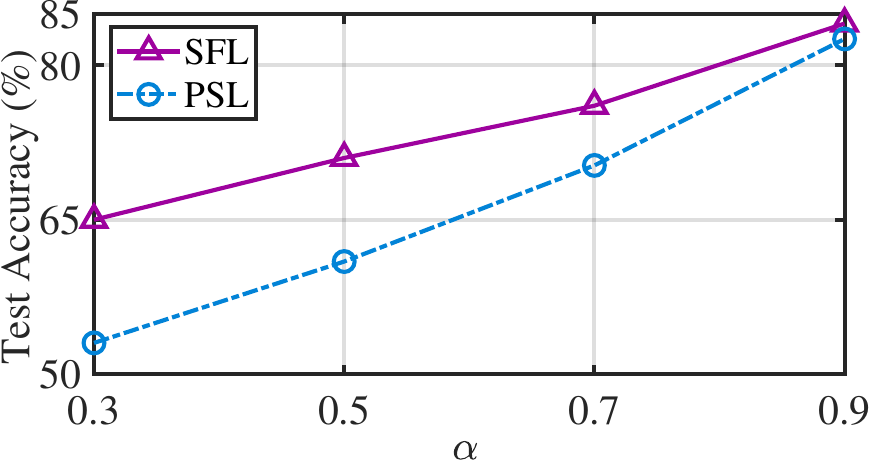}
  }
  \subfloat[Angular deviation.\label{subfig:motivate_2_2}]
  {
    \includegraphics[width=0.238\textwidth]{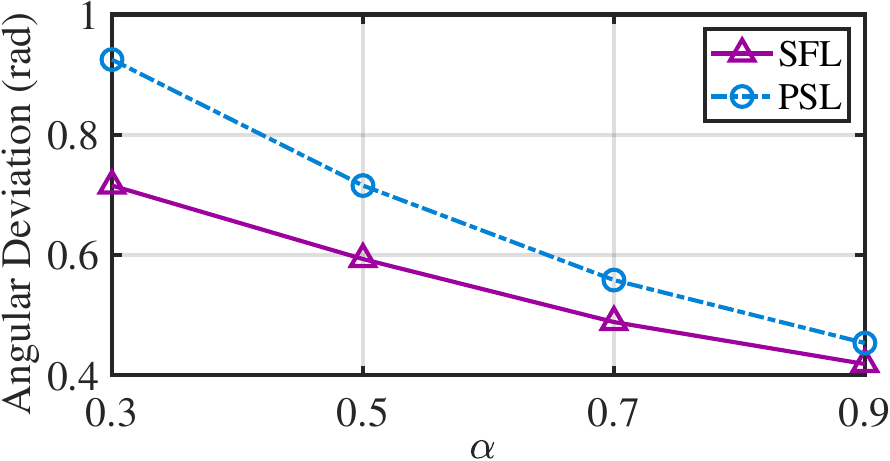}
  }
\vspace{-0.1em}
  \caption{The test accuracy (a) and average angular deviation (b) versus degree of data heterogeneity on CIFAR-10 using VGG-16.}
  \label{fig:motivating_2_total}
\end{figure}

\subsection{Gradient Directional Inconsistency} \label{challenge_2}

While Section~\ref{challenge_1} underscores that the lack of client-side model aggregation deteriorates the training performance of PSL, the underlying cause stems from inconsistent gradient directions across edge devices. In PSL, each device independently performs forward propagation (FP) and BP on its local dataset, without any inter-device aggregation or coordination. In the non-IID setting, this independence drives each device to optimize toward a client-specific objective,  resulting in gradients that diverge from the global optimization direction. While some device gradients may align with the global trend, others, biased by skewed data distribution, can contribute high-variance or even conflicting updates, thereby hindering model convergence.


To validate the above intuition, we compare the test accuracy and gradient directional consistency of SFL and PSL under varying degrees of data heterogeneity. Fig.~\ref{subfig:motivate_2_1} illustrates that the test accuracy of PSL is consistently lower than that of SFL, with the performance gap widening notably as data heterogeneity increases (i.e., smaller $\alpha$). This performance degradation aligns closely with the gradient behavior shown in Fig.~\ref{subfig:motivate_2_2}, where we measure gradient directional consistency using the average angular deviation between gradients\footnote{Let ${\bf g}_i$ and ${\bf g}_j$ denote the gradients computed by the $i$-th and $j$-th edge devices on their local datasets. The pairwise angular deviation is defined as $\arccos(\frac{\langle {\bf g}_i, {\bf g}_j \rangle}{\|{\bf g}_i\| \cdot \|{\bf g}_j\|})$. The values reported in Fig.~\ref{subfig:motivate_2_2} plot the average of these deviations across all distinct device pairs, quantifying the overall directional inconsistency.} (i.e., a smaller angular deviation indicates more consistent gradient directions). The results show that PSL exhibits significantly higher angular deviation than SFL, particularly under high data heterogeneity, indicating more severe gradient misalignment due to the lack of client-side coordination. Taken together, these results reveal a strong correlation between degraded performance and gradient directional inconsistency in PSL. As device gradients become increasingly misaligned, the global model struggles to converge effectively. This provides compelling empirical evidence that gradient directional inconsistency is a key bottleneck for PSL under the non-IID setting.

\section{System Design}\label{sec:s3l_frame}

\subsection{Overview}\label{subsec:preli_overview}

\begin{figure}[t]
\centering
\includegraphics[width=.9\columnwidth]{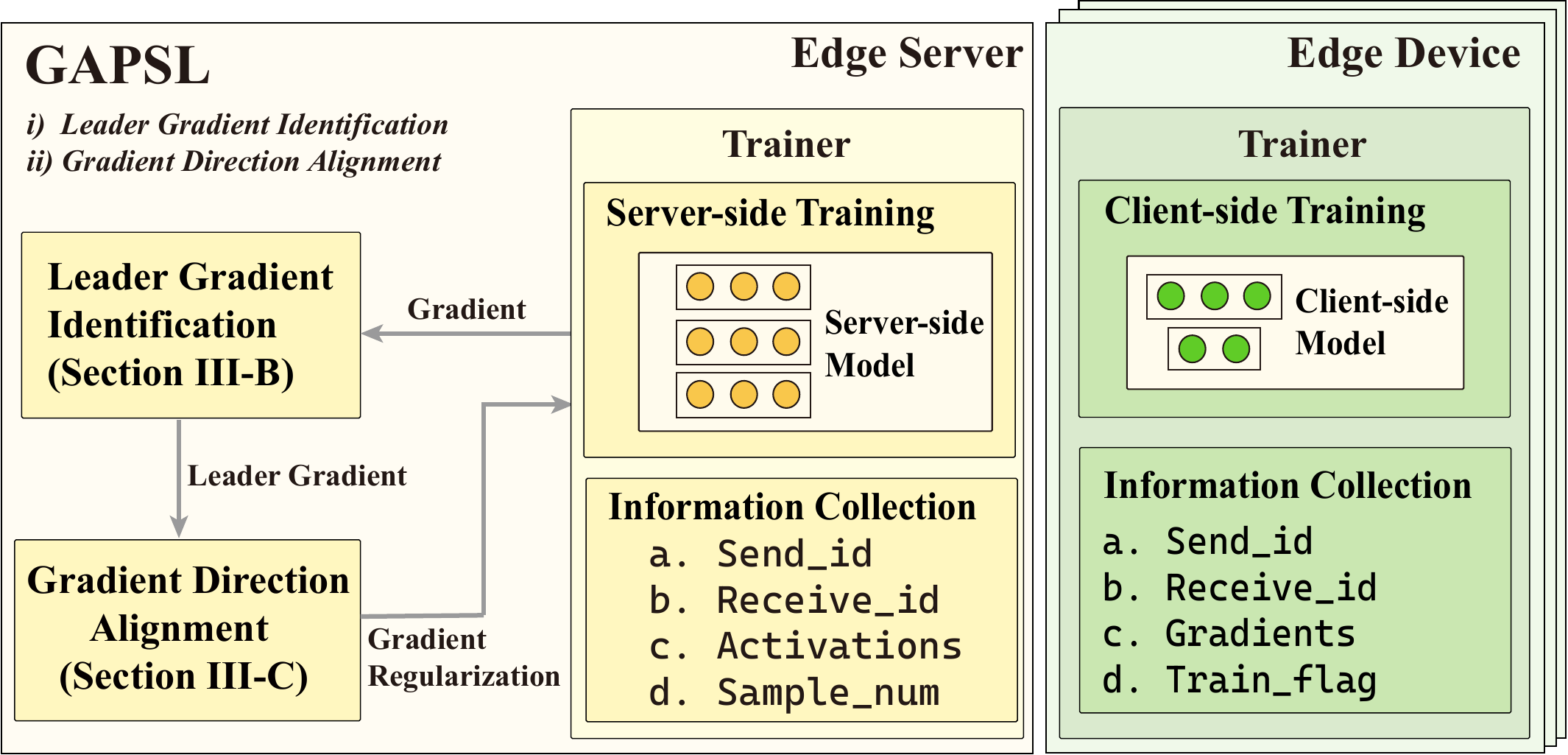}
\vspace{-0.1em}
\caption{
An overview of \name architecture. }
\vspace{-1.0em}
\label{fig:gapsl}
\end{figure}

In this section, we propose \name, a gradient-aligned PSL framework designed to enable efficient and stable model training without client-side model aggregation. 
As illustrated in Fig.~\ref{fig:gapsl}, \name consists of two key components: i) \textit{leader gradient identification} (LGI) (Section~\ref{subsec:LGI}), which identifies the consensus reference direction by selectively aggregating directionally consistent device gradients; and ii) \textit{gradient direction alignment} (GDA) (Section~\ref{subsec:GDA}), which enforces inter-device gradient consistency via direction-aware regularization.


In \name, the global model is partitioned into two sub-models: a client-side model deployed on edge devices and a server-side model hosted on the server. The training workflow for one training round consists of the following three steps: i) each edge device executes FP on the client-side model with its local dataset and transmits the resulting activations to the server; ii) the server employs LGI to construct a leader gradient and then utilizes GDA to regulate each deivce's gradient direction, followed by BP on the server-side model to generate activation's gradients; and iii) the server sends all generated gradients to the edge device for updating their respective client-side models via BP. Notably, since LGI and GDA are executed entirely on the server, \name incurs the same communication overhead as standard PSL (i.e., exchanging activations/gradients), which is strictly lower than SFL due to the elimination of client-side model parameter transmissions. 

\subsection{Leader Gradient Identification}\label{subsec:LGI}

As discussed in Section~\ref{sec:background_motivation}, the lack of client-side model aggregation in PSL exacerbates the divergence of gradient directions across edge devices. Consequently, the contribution of individual devices to the global model convergence is highly uneven: while some gradients promote model convergence, others may introduce noise or adversarial updates driven by local data biases.  To empirically validate this effect, we conduct an experiment where the model is trained using the gradients from a single edge device only\footnote{In reality, PSL cannot do this because a client cannot access another client's gradients to update its model.}. Fig.~\ref{subfig:desin_com_1_1} demonstrates significant variance in test accuracy across edge devices, and Fig.~\ref{subfig:desin_com_1_2} reveals that the most contributive gradient changes across training epochs.

These observations point to a key insight: \textit{due to gradient direction discrepancies, treating all client updates equally can lead to sub-optimal updates and hinder convergence.} This underscores the need to identify the underlying global convergence trend to guide the training process. Although PSL forgoes model aggregation and thus does not access client-side gradients, the server naturally computes gradients for its own model segment for each client. Since the server takes activations from client-side models as input, these server-side gradients inherently capture training divergence stemming from client-side data and model heterogeneity. Motivated by this, we propose the LGI mechanism, as shown in Fig.~\ref{fig:LGI}. LGI dynamically selects a subset of directionally consistent server-side gradients (i.e., those closely aligned with the global convergence direction) to construct a leader gradient. This leader gradient serves as a directional reference (or anchor) for the subsequent gradient alignment in Section~\ref{subsec:GDA}. The LGI mechanism consists of the following three key components:

\begin{figure}[t]
\vspace{-0.9em}
  \centering
  \subfloat[Test accuracy versus gradient.\label{subfig:desin_com_1_1}]{
    \includegraphics[width=0.23\textwidth]{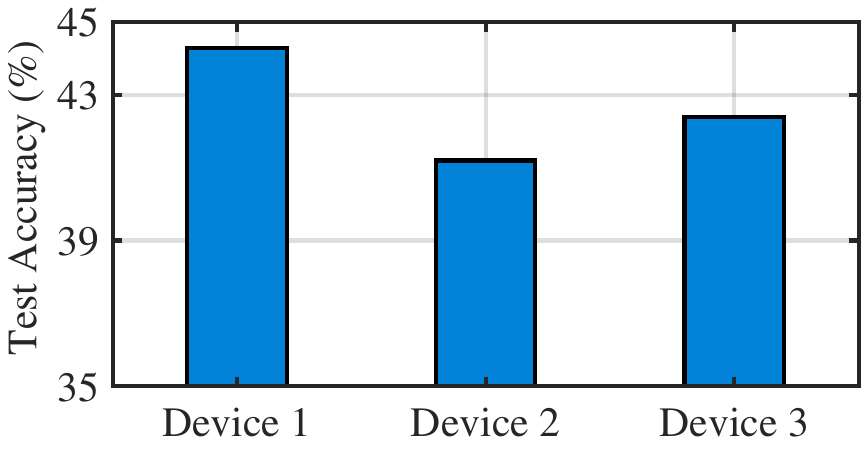}
  }
  \subfloat[Test accuracy versus epochs.\label{subfig:desin_com_1_2}]
  {
    \includegraphics[width=0.23\textwidth]{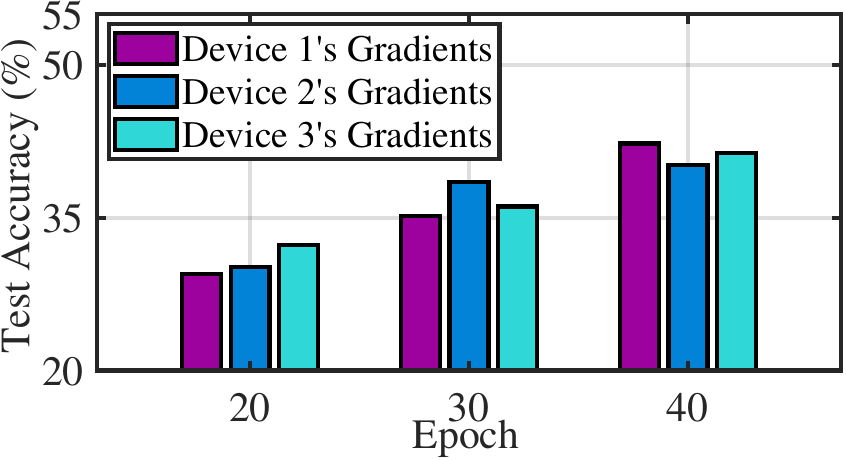}
  }
  \vspace{-0.1em}
  \caption{The PSL performance of training a model with the gradient from a single edge device only (discarding others' gradients) on CIFAR-10 using VGG-16 under the non-IID setting. 
  }
  \label{fig:desin_com_1}
\end{figure}

\subsubsection{Gradient Consistency Scoring}\label{sssec:LGI_1}

Existing approaches for assessing the gradient contribution in distributed ML typically rely on gradient magnitudes~\cite{wang2023aocc,song2019profit,lian2023gofl}, assuming that larger gradients convey more informative updates. However, this assumption often falters in non-IID settings, as large gradients may deviate from the global convergence direction and thus impede model convergence. In contrast, small but directionally consistent gradients are more effective in promoting stable convergence. To rectify this, we propose a gradient consistency score that quantifies the directional alignment of a device’s gradient relative to others. For the $t$-th training round, the gradient consistency score of the $i$-th edge device ${\theta}_{i}^t$ is defined as the average angular deviation between its gradient and those of other participating devices:
\begin{equation}\label{eqn_gradient_consistency_scoring}
{\theta}_i^t = \frac{1}{|\mathcal{S}| - 1} \sum_{\substack{j \in \mathcal{S} \\ j \ne i}} {\theta}_{i, j}^t, \;\text{where}\; {\theta}_{i, j}^t = \arccos\left( \frac{ \langle {\bf g}^t_i, \, {\bf g}^t_j \rangle }{ \|{\bf g}^t_i\| \cdot \|{\bf g}^t_j\| } \right),
\end{equation}
where $\mathcal{S} = \left\{ {1,2,...,S} \right\}$ denotes the set of participating edge devices, ${\theta}_{i, j}^t$ represents the angular deviation between the gradients of $i$-th and $j$-th edge device, $\mathbf{g}_k^t$ is the gradient of the $k$-th edge device for the server-side model, $\arccos\left( \cdot \right)$, ${\langle \cdot,  \cdot \rangle}$, and $\|\cdot\|$ denote the inverse cosine function, the inner product operation, and the Euclidean norm, respectively.

Eqn.~\eqref{eqn_gradient_consistency_scoring} yields two key insights regarding the geometric properties of client updates: i)  A smaller ${\theta}_i^t$ indicates that the client’s gradient is directionally consistent with the majority, implying a higher likelihood of constructive contribution to the global optimization; and ii) A larger ${\theta}_i^t$ indicates severe directional inconsistency, suggesting that the client’s gradient may introduce noise or conflicting updates. As illustrated in Fig.~\ref{fig:desin_com_2}, the edge device with a lower gradient consistency score consistently yields superior test accuracy. This result reveals the effectiveness of the proposed consistency scoring in identifying directionally consistent gradients.


\subsubsection{Adaptive Gradient Selection}\label{sssec:LGI_2}

To mitigate the adverse effects of conflicting updates, we propose an adaptive gradient selection strategy that prioritizes gradients with stronger directional consistency (i.e., those with lower gradient consistency scores as defined in Eqn.~\eqref{eqn_gradient_consistency_scoring}). These selected gradients form the basis for constructing the leader gradient, which serves as a directional anchor for guiding the model training (See Section~\ref{sssec:LGI_3}). While gradient consistency scoring helps quantify the quality of individual gradients, a key design question arises: \textit{how many gradients should be selected?}

 \begin{figure}[t]
\centering
\includegraphics[width=.90\columnwidth]{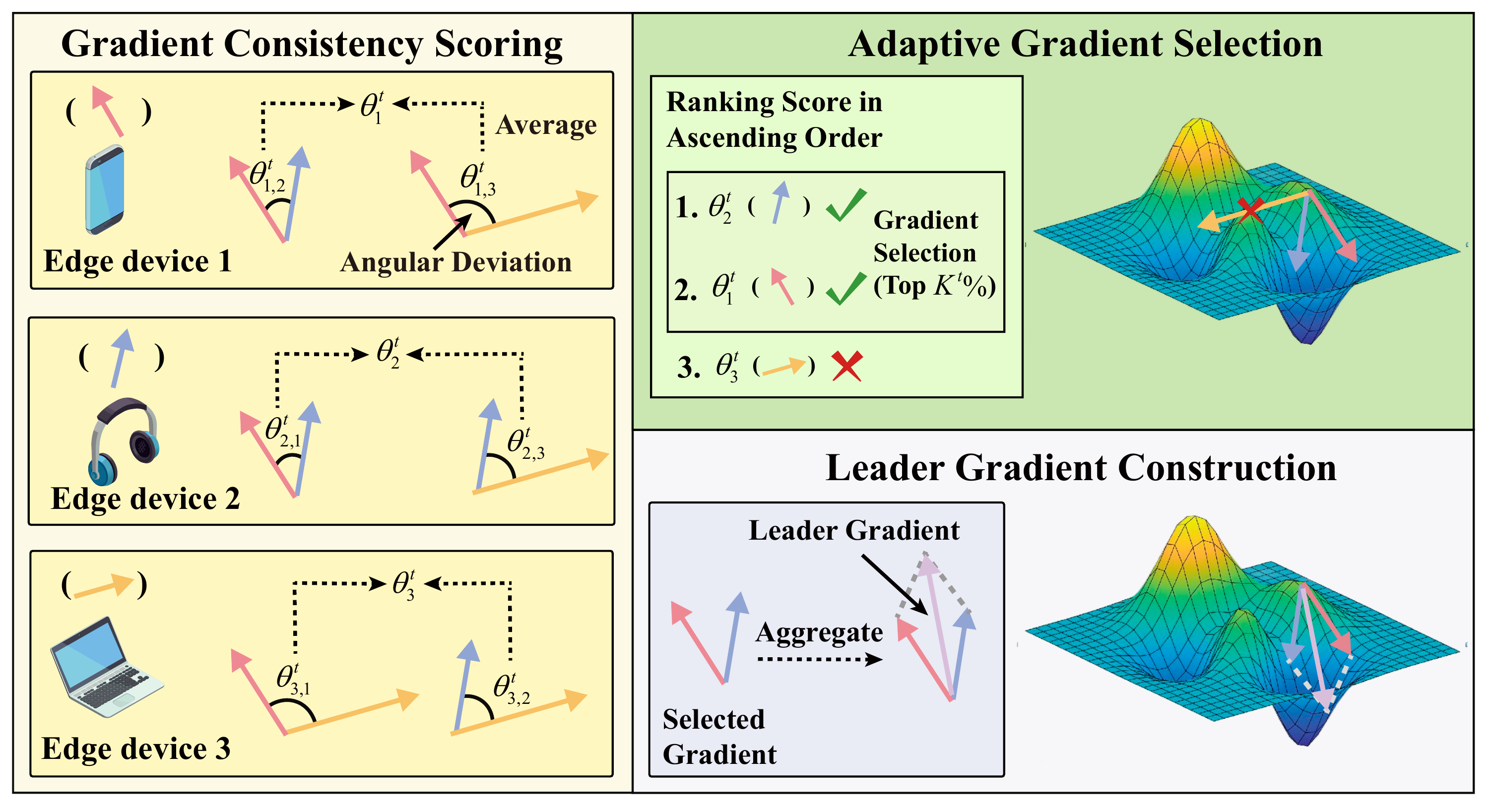}
\vspace{-0.3em}
\caption{ An illustration of LGI. }
\label{fig:LGI}
\end{figure}

\begin{figure}[t]
  \centering
   \vspace{-2.3em}
  \subfloat[Normalized consistency score.\label{subfig:desin_com_2_1}]{
    \includegraphics[width=0.23\textwidth]{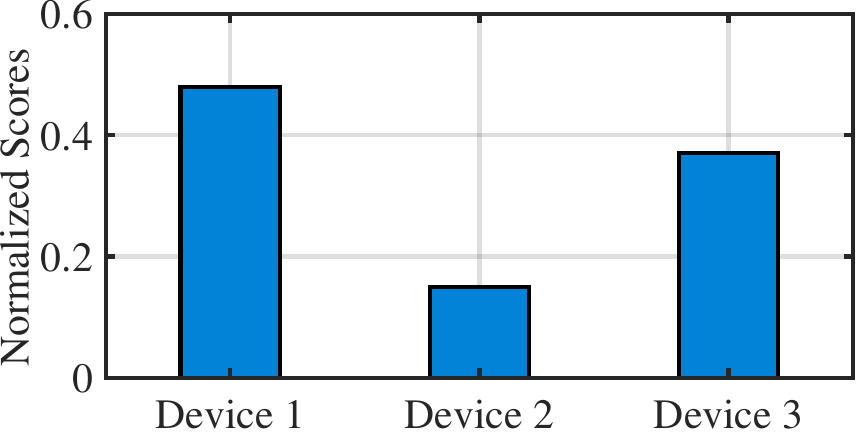}
  }
  \subfloat[Test accuracy.\label{subfig:desin_com_2_2}]
  {
    \includegraphics[width=0.23\textwidth]{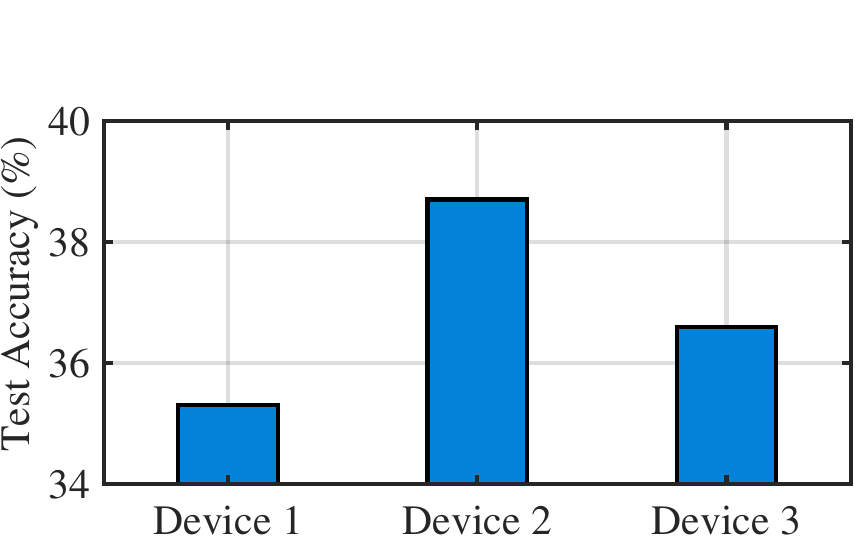}
  }
    \vspace{-0.1em}
  \caption{The normalized gradient consistency scores and corresponding test accuracies of edge devices over 30 epochs on CIFAR-10 using VGG-16.}
  \label{fig:desin_com_2}
\end{figure}

\begin{figure}[t]
\vspace{-1.1em}
  \centering
  \subfloat[Test accuracy versus selection ratio.\label{subfig:desin_com_3_1}]{
    \includegraphics[width=0.23\textwidth]{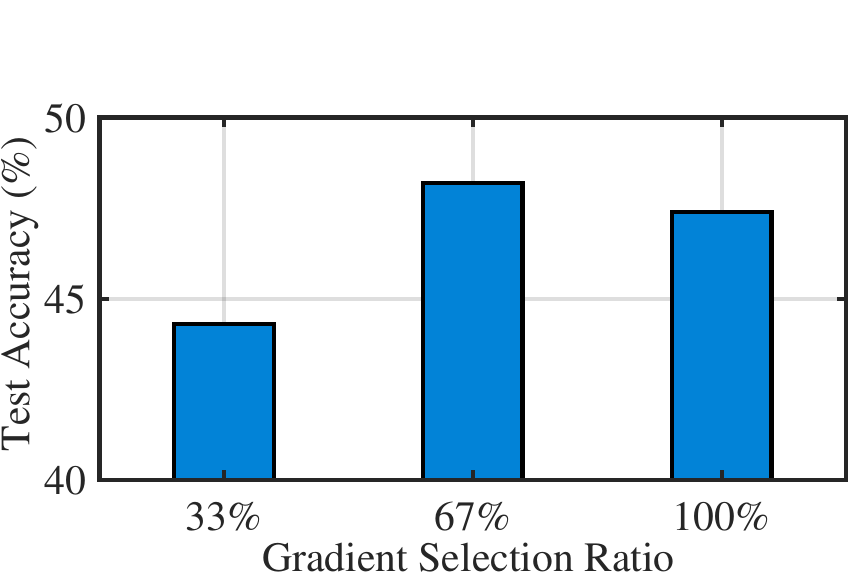}
  }
  \subfloat[Test accuracy versus epochs.\label{subfig:desin_com_3_2}]
  {
    \includegraphics[width=0.23\textwidth]{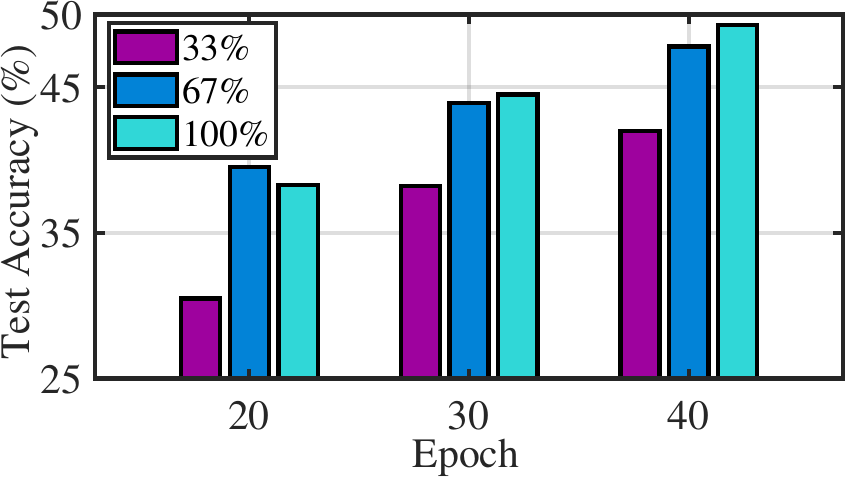}
  }
    \vspace{-0.1em}
  \caption{The test accuracy versus gradient selection ratio and epochs on CIFAR-10 using VGG-16.}
  \label{fig:desin_com_3}
\end{figure}

We introduce an adaptive gradient selection ratio $K\%$, defined as the time-varying proportion of top-ranked gradients (i.e., sort gradients in ascending order of consistency scores) selected. This ratio governs a fundamental trade-off between gradient directional alignment and information diversity: selecting too few gradients enhances gradient consistency but limits representational diversity, while selecting too many increases the risk of incorporating divergent or noisy gradients. This theoretical trade-off is empirically corroborated by Fig.~\ref{subfig:desin_com_3_1}, which demonstrates that extremes in selection ratios degrade model performance. Furthermore, Fig.~\ref{subfig:desin_com_3_2} reveals that the optimal selection ratio is not static but varies across training epochs. 
These observations underscore the inadequacy of fixed selection ratios and highlight the necessity of an adaptive mechanism capable of adjusting the selection ratio for training dynamics.

\begin{figure}[t]
\centering
\includegraphics[width=.82\columnwidth]{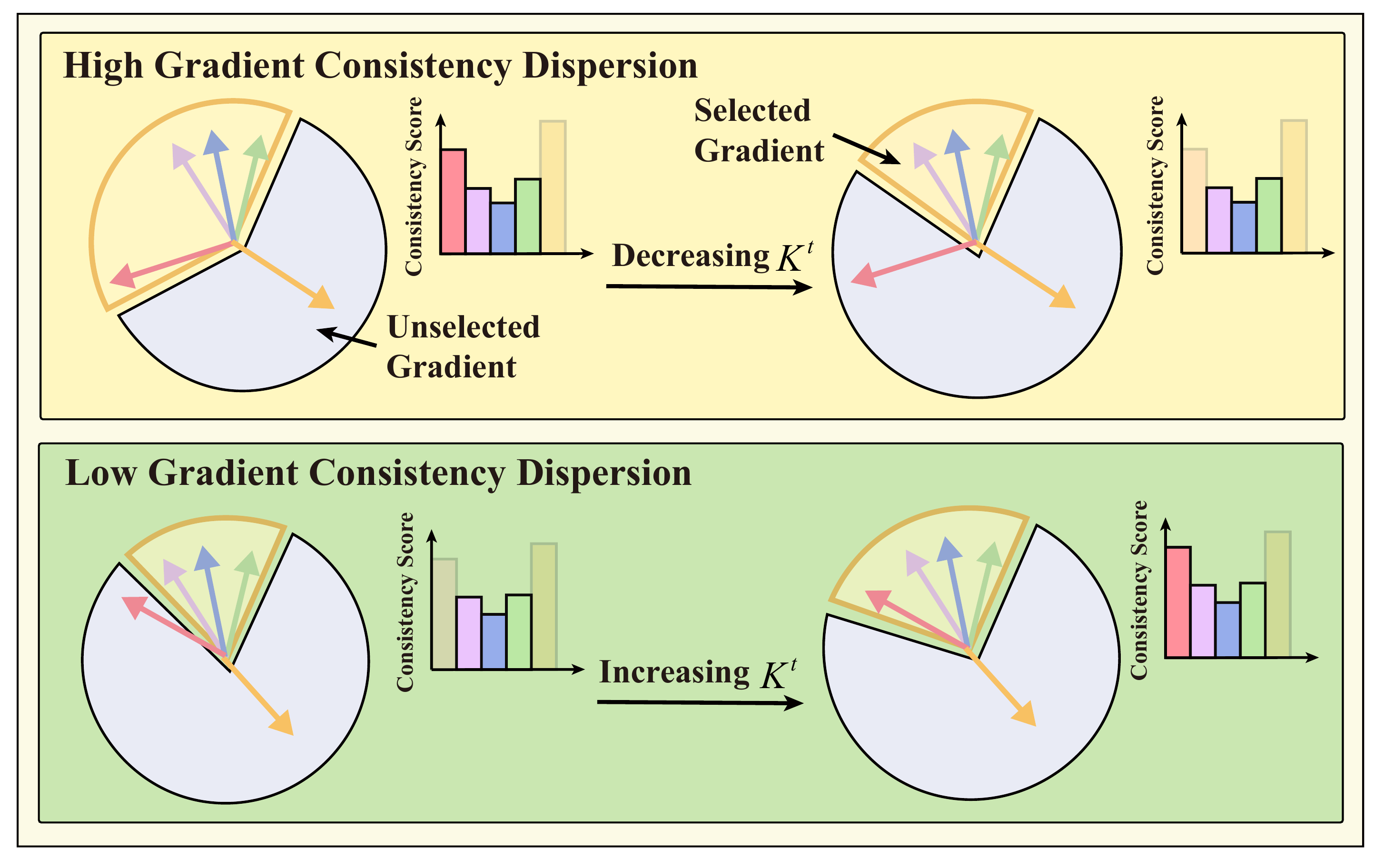}
  \vspace{-0.3em}
\caption{ The impact of gradient consistency dispersion on gradient selection.}
\label{fig:variance_gradient_seclect}
\end{figure}

\begin{figure}[t]
  \centering
  \vspace{-0.0em}
  \subfloat[Test accuracy versus selection ratio.\label{subfig:desin_com_4_1}]{
    \includegraphics[width=0.23\textwidth]{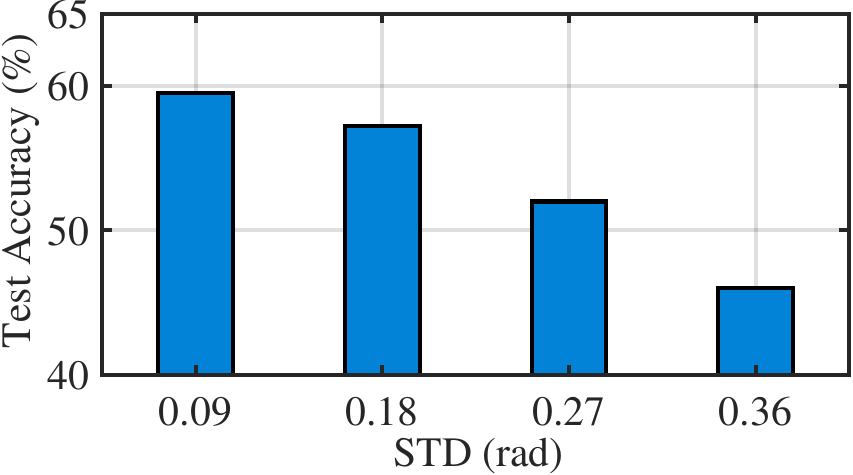}
  }
  \subfloat[Test accuracy versus epochs.\label{subfig:desin_com_4_2}]
  {
    \includegraphics[width=0.23\textwidth]{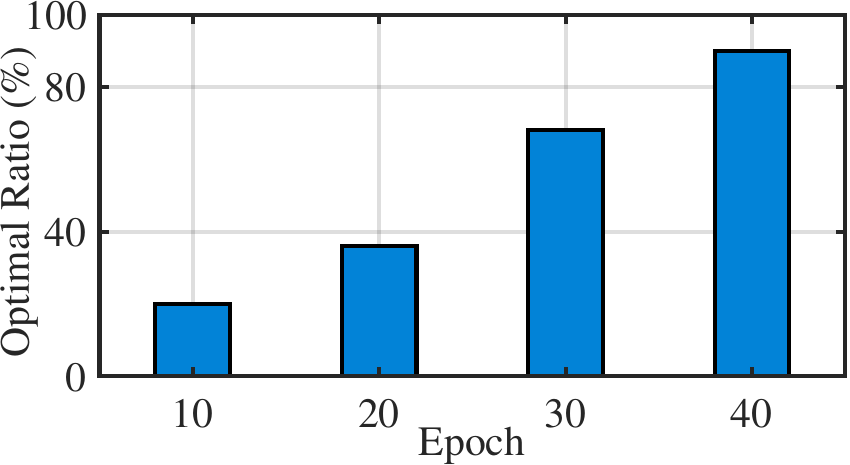}
  }
    \vspace{-0.1em}
  \caption{The test accuracy versus STD of gradient consistency score and optimal selection ratio versus epochs on CIFAR-10 using VGG-16.
  }
  \label{fig:desin_com_4}
\end{figure}

To design an adaptive gradient selection mechanism, we begin with two intuitions: i) \textit{Gradient consistency dispersion}: The dispersion of gradient consistency scores serves as an indicator of directional consensus among device gradients. As shown in Fig.~\ref{fig:variance_gradient_seclect}, a high dispersion implies substantial disagreement in gradient directions; a stringent selection strategy is imperative to filter out conflicting updates and stabilize training.  Conversely, low dispersion suggests a tight clustering of gradient directions, allowing for the inclusion of mid-ranked gradients to enhance information utilization without compromising training performance; ii) \textit{Training dynamics}: The optimal gradient selection ratio is not static but evolves throughout the training process. Crucially, rather than requiring exhaustive task-specific manual tuning that would introduce severe hyperparameter optimization overhead, the gradient selection ratio is designed to dynamically self-adjust. During the early stages, a conservative selection ratio is essential to enforce strict directional alignment and suppress noisy gradients. As the model converges, progressively expanding the selection to include a broader range of gradients introduces beneficial diversity and thus improves model generalization. To validate these two design principles, we conduct two empirical studies using the standard deviation (STD) of gradient consistency scores to quantify gradient directional dispersion\footnote{The gradient consistency score captures the angular deviation of each client’s gradient relative to the others, reflecting how well it aligns with the collective update direction. The STD of these scores indicates the degree of directional agreement: a small STD indicates that most device gradients deviate by similar angles, suggesting strong gradient directional alignment among clients; in contrast, a large STD implies significant variation, with some gradients closely aligned and others highly deviated.}. Fig.~\ref{subfig:desin_com_4_1} illustrates that test accuracy exhibits a consistent decline as the STD increases, indicating that higher dispersion undermines training performance. Furthermore, an exhaustive search for the optimal selection ratio across training epochs, as shown in Fig.~\ref{subfig:desin_com_4_2}, reveals a monotonic increase in the optimal selection ratio. This trend confirms the necessity of a time-varying selection that transitions from strict alignment to broader participation as the global model converges.


The design of our adaptive selection ratio is built on the empirical insights observed in Fig.~\ref{fig:desin_com_4}. The negative correlation between gradient dispersion and training performance (Fig.~\ref{subfig:desin_com_4_1}) mandates a stability-aware penalty mechanism. We encode this by normalizing the current dispersion $\nu({\boldsymbol{\theta}}^t)$ against its historical maximum to derive a relative stability score, which adaptively constricts the selection ratio as directional divergence intensifies. Temporally, the linear growth of the optimal selection ratio (Fig.~\ref{subfig:desin_com_4_2}) motivates the incorporation of a linear time-scaling term, $\frac{t}{T}$, designed to progressively relax selection constraints as the model converges. We employ a multiplicative coupling of these two factors that functions as a robust gating mechanism: it ensures that the selection ratio approaches its upper bound $K_{max}$ if and only if the training is sufficiently advanced and the gradients remain directionally consistent. This design explicitly prevents the inclusion of conflicting updates for high directional divergence. Therefore, the selection ratio at the $t$-th training round is defined as:
\begin{equation}\label{eqn_K_t}
{K^t} = {K_{\min }} + \left( {\frac{t}{T}\frac{{ {\nu^t_{\max }-\nu ({\boldsymbol{\theta}}^t)}}}{{{\nu^t_{\max }} - {\nu^t_{\min }}}}} \right)({K_{\max }} - {K_{\min }}),
\end{equation}
where ${{\boldsymbol{\theta}}^t}  = \{ \theta^t_i \}_{i=1}^{S}$ is the set of gradient consistency scores for all edge devices in the $t$-th training round; ${\nu}(\boldsymbol{\theta})$ represents the STD of directional consistency scores, quantifying the dispersion of gradient directions; ${K_{\min }}$ and ${K_{\max}}$ are lower and upper bounds of the feasible selection ratio range\footnote{${K_{\min }}$ and ${K_{\max}}$ can be pre-set based on data heterogeneity, e.g., smaller values are preferred for highly non-IID datasets to reduce conflicting updates, while larger values suit more homogeneous datasets to speed up convergence.}, respectively; ${\frac{{ {\nu^t_{\max }-\nu ({\boldsymbol{\theta}}^t)}}}{{{\nu^t_{\max }} - {\nu^t_{\min }}}}}$ functions as the relative stability score, which maps the current gradient volatility into a normalized range $[0, 1]$, where $\nu^t_{\min}$ and $\nu^t_{\max}$ are dynamically updated via tracking the minimum and maximum observed values of $\nu(\boldsymbol{\theta}^t)$ observed up to the $t$-th round.

{\bf{Remark.}} In Eqn.~\eqref{eqn_K_t}, the term $\nu({\boldsymbol{\theta}}^t)$ serves as a key indicator for adaptive adjustment of the selection ratio $K^t$: when dispersion is high (i.e., gradients are misaligned), $K^t$ is reduced to exclude potentially conflicting updates; when dispersion is low (i.e., gradients are well-aligned), a larger $K^t$ is enabled to incorporate more client contributions.  In addition, the temporal scaling term $\frac{t}{T}$ ($T$ denotes the pre-scheduled total number of training rounds, requiring no additional task-specific tuning) introduces a progressive increase in $K^t$ over the model training. This design encourages more conservative updates during the early stages to prioritize gradient stability and allows for broader device gradient participation in later stages, which improves model generalization as training converges.

\subsubsection{ Leader Gradient Construction} \label{sssec:LGI_3}

\RestyleAlgo{ruled}
\LinesNumbered
\SetKwFunction{Fns}{Gradient Consistency Scoring}
\SetKwProg{Fn}{}{:}{}
\SetKwFunction{Fu}{Adaptive Gradient Selection}
\SetKwFunction{Fc}{Leader Gradient Construction}
\SetKwInOut{Output}{Output}
\begin{algorithm}[t]
\caption{Leader Gradient Identification}\label{alg:LGI}
\KwIn{
    $\mathcal{S}$, $K_{\min}, K_{\max}$, $\nu_{\min}^t$, $\nu_{\max}^t$, $t$, and $T$
}
\KwOut{${\bf g}_{\text{lead}}^t$}
\BlankLine
\Fn{\Fns}{
    \ForEach{$i \in \mathcal{S}$}{
        ${\theta}_i^t = \frac{1}{|\mathcal{S}| - 1} \sum_{\substack{j \in \mathcal{S} \\ j \ne i}} \arccos\left( \frac{ \langle {\bf g}_i, \, {\bf g}_j \rangle }{ \|{\bf g}_i\| \cdot \|{\bf g}_j\| } \right)$ ~~~// \textit{Compute Angular
Deviation}\;  
    }
    ${{\boldsymbol{\theta}}^t}  \gets \{ \theta^t_i \}_{i=1}^{S}$\;
}
\Fn{\Fu}{
    $\nu({\boldsymbol{\theta}}^t) \gets \operatorname{Std}({\boldsymbol{\theta}}^t)$ ~~~// \textit{Compute Standard Deviation}\;
    $\nu_{\min}^t \gets \min(\nu_{\min}^{t-1}, \nu(\boldsymbol{\theta}^t))$\;
    $\nu_{\max}^t \gets \max(\nu_{\max}^{t-1}, \nu(\boldsymbol{\theta}^t))$\;
    ${K^t} = {K_{\min }} + \left( {\frac{t}{T}\frac{{ {\nu^t_{\max }-\nu ({\boldsymbol{\theta}}^t)}}}{{{\nu^t_{\max }} - {\nu^t_{\min }}}}} \right)({K_{\max }} - {K_{\min }})$ ~// \textit{Calculate Gradient Selection Ratio}\;
}

\Fn{\Fc}{
    Sort $\mathcal{S}$ by $D_i^t$ in ascending order\;
    $\mathcal{{\widetilde S}}^t = \left\{ i \in \mathcal{{ S}} \;\middle|\; \pi(\theta_i^t) \leq \lceil K^t\% \cdot |\mathcal{S}| \rceil \right\}$ ~// \textit{Select Top-$K^t\%$ edge devices}\;
  $\mathcal{\widetilde G}^t \gets \{ {\bf{g}}_i^t \mid i \in \mathcal{{\widetilde S}}^t \}$\;
  ${\bf g}_{\text{lead}}^t \gets \frac{1}{|\mathcal{\widetilde G}^t|} \sum_{{\bf g}_i^t \in \mathcal{\widetilde G}^t} {\bf g}_i^t$ ~// \textit{Construct Leader Gradient}\;
}
\Return{${\bf g}_{\mathrm{lead}}^t$}
\end{algorithm}

With the adaptive selection ratio $K^t$ established, the final step of the LGI is to construct a leader gradient, a robust estimate of the global optimization direction that serves as a directional anchor for gradient alignment in Section~\ref{subsec:GDA}. We first rank all device gradients in ascending order of their consistency scores, such that lower scores (indicating greater alignment) are ranked higher. Then, we select the top 
$\lceil K^t\% \cdot |\mathcal{S}| \rceil$ edge devices to form the gradient subset, as given by 
\begin{equation}
\mathcal{{\widetilde S}}^t = \left\{ i \in \mathcal{{ S}} \;\middle|\; \pi(\theta_i^t) \leq \lceil K^t\% \cdot |\mathcal{S}| \rceil \right\},
\end{equation}
where $\pi(\theta_i^t)$ denotes the ascending rank index of device $i$'s consistency score (i.e., lower scores imply higher consistency). The leader gradient is then computed via the selective aggregation of the gradient subset $\mathcal{\widetilde G}^t = \{ {\bf{g}}_i^t \mid i \in \mathcal{{\widetilde S}}^t \}$: 

\begin{equation}
    {\bf g}_{\text{lead}}^t = \frac{1}{|\mathcal{\widetilde G}^t|} \sum_{{\bf g}_i^t \in \mathcal{\widetilde G}^t} {\bf g}_i^t.
\end{equation}

By exclusively incorporating directionally consistent updates, this construction effectively filters out statistical outliers and adversarial noise. As a result, the leader gradient provides a noise-resilient and directionally reliable proxy for the true global descent direction. The complete LGI procedure is summarized in {\bf{Algorithm~\ref{alg:LGI}}}.

\begin{figure}[t]
 \vspace{-0.8em}
  \centering
  \subfloat[Test accuracy.\label{subfig:desin_com_5_1}]{
    \includegraphics[width=0.233\textwidth]{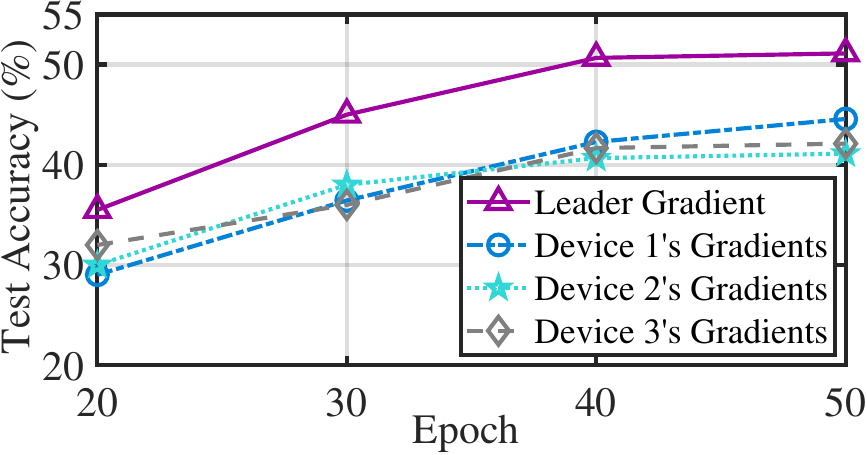}
  }
  \subfloat[Converged time.\label{subfig:desin_com_5_2}]
  {
    \includegraphics[width=0.225\textwidth]{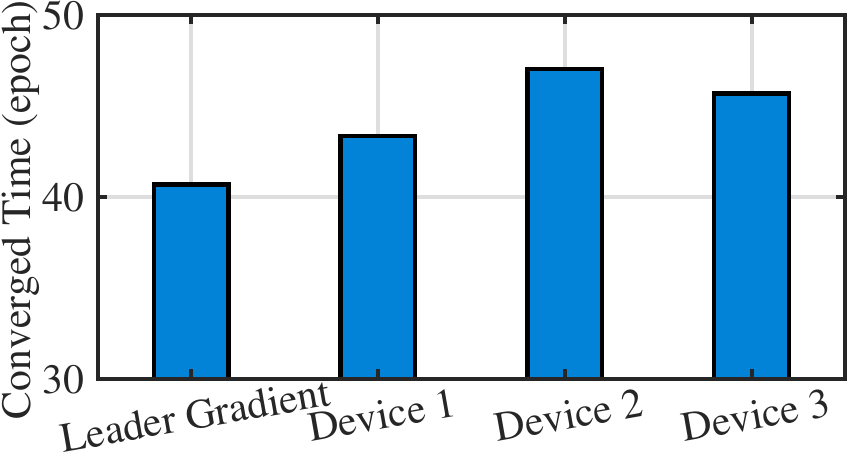}
  }
    \vspace{-0.1em}
  \caption{The test accuracy and converged time via leader gradient and individual gradients on CIFAR-10 using VGG-16.}
  \label{fig:desin_com_5}
\end{figure}

\subsection{Gradient Direction Alignment}\label{subsec:GDA}

Following the construction of the leader gradient via LGI (Section~\ref{subsec:LGI}), we obtain a directional anchor that serves as a robust proxy for the global convergence trend. Empirical validation in Fig.~\ref{fig:desin_com_5} demonstrates that models updated with this leader gradient significantly outperform those relying on individual gradients in test accuracy and convergence speed. These results underscore the potential of the leader gradient to guide or correct device gradient directions for enhancing training performance.



\begin{figure}[t]
\centering
\includegraphics[width=.91\columnwidth]{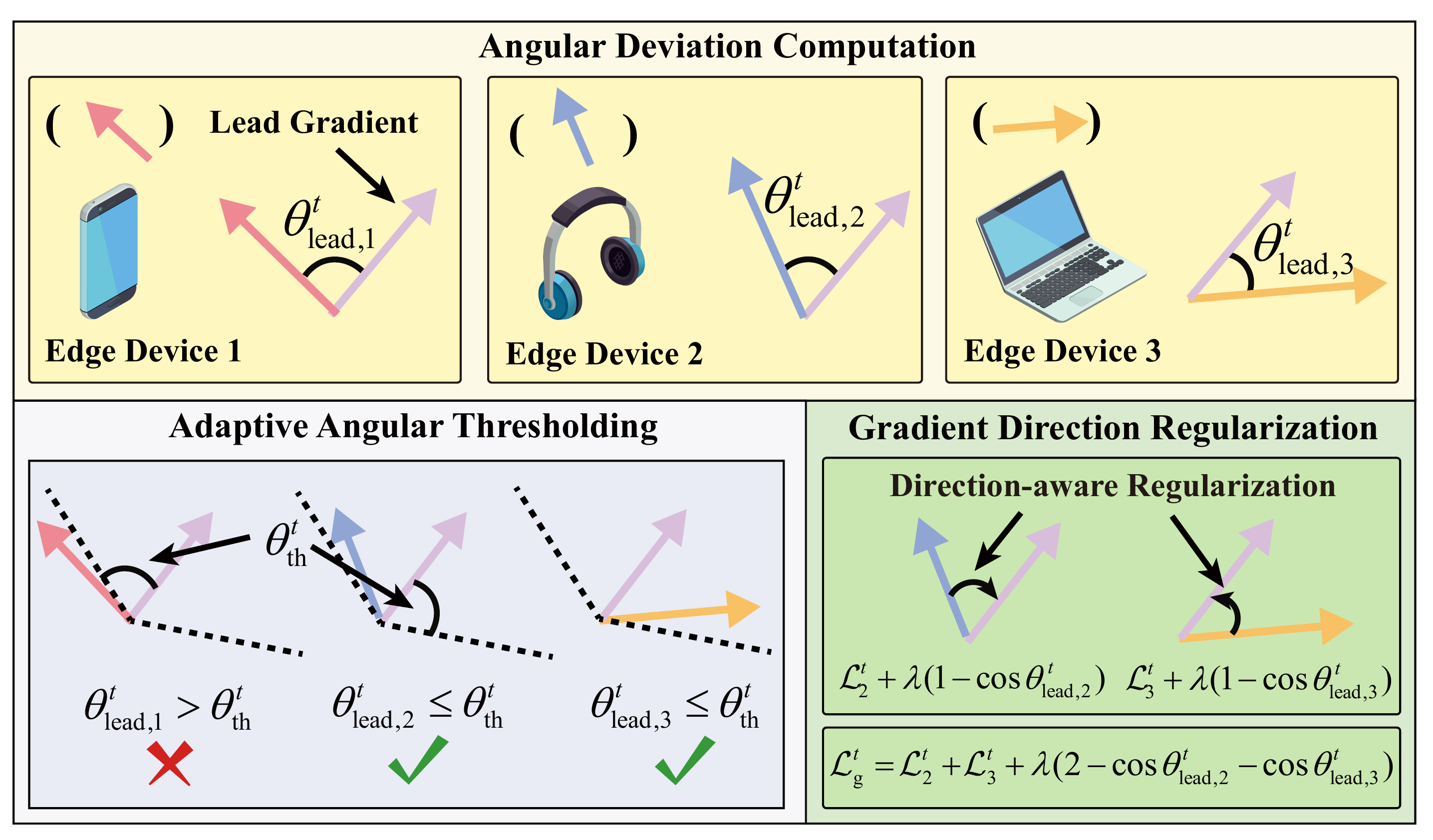}
  \vspace{-0.3em}
\caption{ An illustration of GDA. }
\label{fig:GDA}
\end{figure}

In light of this, we propose the GDA mechanism, as illustrated in Fig~\ref{fig:GDA}, to rectify the optimization drift induced by data heterogeneity. Rather than naively forcing all local updates to align with the leader gradient—which risks model degradation due to extreme statistical outliers—GDA adopts a robust two-stage alignment strategy. First, it employs an adaptive angular threshold to identify and exclude severely divergent gradient updates. Subsequently, it applies direction-aware regularization exclusively to the remaining viable clients. This selective alignment ensures that the global model assimilates beneficial local representations while strictly isolating destructive noise. Specifically, the GDA framework comprises the following three key components:


\subsubsection{Angular Deviation  Computation} \label{sssec:GDA_1}

To quantify the alignment between a device's local update and the consensus reference direction, GDA computes the angular deviation between each device’s gradient and the leader gradient ${\bf g}_{\text{lead}}$ (derived via LGI in Section~\ref{subsec:LGI}). For the $t$-th training round, the angular deviation between the device $i$' gradient and the leader gradient is calculated as 
\begin{equation}\label{eqn:theta}
\theta_{{\text{lead}}, i}^t = \arccos\left( \frac{\langle \mathbf{g}^t_i, \mathbf{g}^t_{\text{lead}} \rangle}{\|\mathbf{g}^t_i\| \cdot \|\mathbf{g}^t_{\text{lead}}\|} \right).
\end{equation}

{\bf{Remark.}} This angular metric serves as a geometric indicator of gradient alignment: A small $\theta_{{\text{lead}}, i}$ (approaching 0) indicates that the device's gradient is well-aligned and directionally consistent with the global convergence direction. Conversely, a large $\theta_{{\text{lead}}, i}$ (beyond $\frac{\pi}{2}$) suggests a conflicting update direction, stemming from severe data heterogeneity or noise. This angular metric provides a principled geometric basis for the subsequent coordination mechanisms, including adaptive thresholding and gradient direction regularization.

\subsubsection{Adaptive Angular Thresholding}\label{sssec:GDA_2}

\begin{figure}[t]
  \centering
  \vspace{-1em}
  \subfloat[Test accuracy versus angular threshold.\label{subfig:desin_com_6_1}]{
    \includegraphics[width=0.235\textwidth]{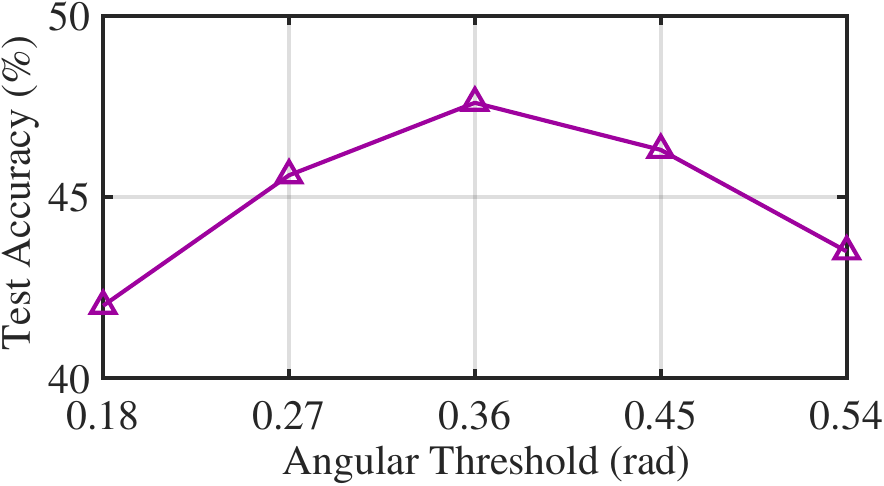}
  }
  \subfloat[Test accuracy versus epochs.\label{subfig:desin_com_6_2}]
  {
    \includegraphics[width=0.23\textwidth]{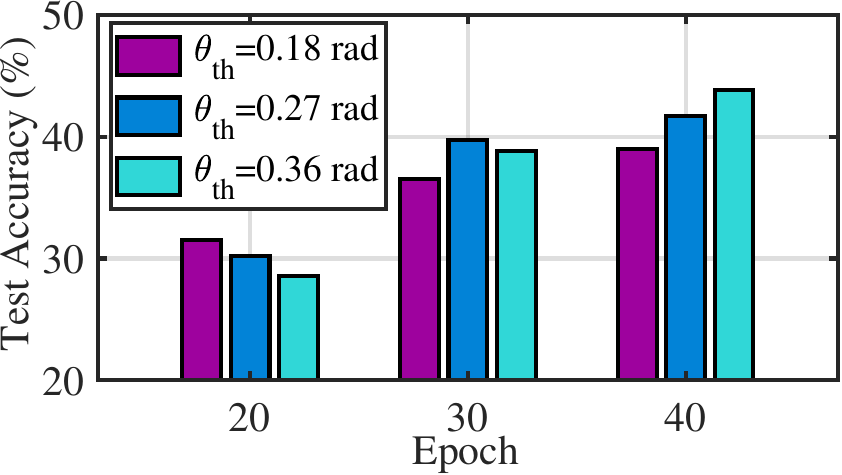}
  }
    \vspace{-0.1em}
  \caption{The test accuracy versus angular threshold and epochs on CIFAR-10 using VGG-16.}
  \label{fig:desin_com_6}
\end{figure}

As discussed in Section~\ref{challenge_2}, device gradients of PSL often exhibit significant directional divergence under non-IID data. Directly aggregating all device gradients, regardless of their alignment with the global optimization direction, may introduce conflicting updates that exacerbate optimization drift and impede model convergence. Consequently, filtering gradients with excessive angular deviations is a prerequisite for effective alignment. 
While employing a fixed angular threshold $\theta_{\text{th}}$ offers a straightforward solution, allowing only device gradients whose angular deviation satisfies $\theta_{{\text{lead}}, i}^t \le \theta_{{\text{th}}}$ to participate in the alignment process, it is inherently rigid and fails to accommodate the evolving gradient dynamics. Fig.~\ref{fig:desin_com_6} reveals the limitations of fixed thresholding. Fig.~\ref{subfig:desin_com_6_1} indicates that extreme thresholds yield sub-optimal performance: overly stringent thresholds discard informative gradients, while overly loose ones admit conflicting noise. Furthermore, Fig.~\ref{subfig:desin_com_6_2} reveals that the optimal threshold changes as training progresses, underscoring the necessity of a dynamic angular threshold. 

To this end, we propose an adaptive angular thresholding mechanism. Since gradient consistency is relative to the distribution of the device gradients for the current training round, using fixed or historical standards is ineffective for distinguishing valid updates from noise. For instance, a larger angular deviation might be acceptable in a high-variance scenario but considered an outlier when gradients are well-aligned. Therefore, effective filtering requires a distribution-based approach that adapts to the current distribution. We avoid using historical min-max scaling because it is sensitive to outliers, i.e., a single gradient with an extreme angle can significantly skew the normalization range, rendering the filter ineffective. Instead, we formulate the threshold by leveraging the first- and second-order statistics of the current gradient distribution to construct a robust confidence boundary. In the $t$-th training round, the angular threshold is derived as:
\begin{equation}\label{eqn:threshold}
\theta _{{\rm{th}}}^t = \max \left( \min \left( {\mu ({\bf{{\boldsymbol{\theta}} }}_{{\rm{lead}}}^t) - \eta \nu ({\bf{{\boldsymbol{\theta}}}}_{{\rm{lead}}}^t),\frac{\pi }{2}} \right), 0 \right),
\end{equation}
where ${{\boldsymbol{\theta}}_{\text{lead}}^t}  = \{ \theta^t_{{\text{lead}},  i} \}_{i=1}^{S}$ is the set of gradient angular deviation for all devices in the $t$-th round, $\mu(\cdot)$ and $\nu(\cdot)$ denote the mean and standard deviation operations, respectively, and $\eta$ controls the sensitivity of the threshold to gradient direction dispersion. Specifically, a larger $\eta$ imposes a more conservative threshold, aggressively penalizing and discarding gradients under high directional variance. The threshold is capped at $\frac{\pi }{2}$ to exclude gradients with opposite directions, which could destabilize model training. Consequently, only clients satisfying $\theta_{\text{lead}, i}^t \leq \theta_{\text{th}}^t$ are selected for the subsequent gradient direction alignment (Section~\ref{sssec:GDA_3}).

{\bf{Remark.}} The adaptive angular threshold in Eqn.~\eqref{eqn:threshold} provides a dynamic adaptation based on the statistical characteristics of device gradients. The mean $\mu(\boldsymbol{\theta}_{\text{lead}}^t)$ serves as the central reference, while the standard deviation $\nu(\boldsymbol{\theta}_{\text{lead}}^t)$ acts as a penalty for gradient directional inconsistency. Specifically, when gradient directions are well-aligned (i.e., low dispersion), the threshold is relaxed to allow broader client participation to accelerate convergence. Conversely, when gradient dispersion is high, the threshold is automatically tightened to enforce a stricter filter against noise.

\RestyleAlgo{ruled}
\LinesNumbered
\SetKwFunction{Fgda}{Gradient Direction Alignment}
\SetKwProg{Fn}{}{:}{}
\SetKwFunction{Fd}{Angular Deviation Computation}
\SetKwFunction{Fth}{Adaptive Thresholding}
\SetKwFunction{Fr}{Gradient Direction Regularization}
\SetKwInOut{Input}{Input}
\SetKwInOut{Output}{Output}
\begin{algorithm}[t]
\caption{Gradient Direction Alignment}\label{alg:GDA}
\Input{
    $\mathcal{S}$, ${\bf g}_{\text{lead}}^t$, $\lambda$
}
\Output{$\mathcal{L}^t_{\text{g}}$}
\BlankLine

\Fn{\Fd}{
    \ForEach{$i \in \mathcal{S}$}{
        $\theta_{{\text{lead}}, i}^t = \arccos\left( \frac{\langle \mathbf{g}^t_i, \mathbf{g}^t_{\text{lead}} \rangle}{\|\mathbf{g}^t_i\| \cdot \|\mathbf{g}^t_{\text{lead}}\|} \right)$ ~~~// \textit{Angular Deviation}\;
    }
    ${{\boldsymbol{\theta}}_{\text{lead}}^t}  = \{ \theta^t_{{\text{lead}},  i} \}_{i=1}^{S}$\;
}

\Fn{\Fth}{
    $\mu({{\boldsymbol{\theta}}_{\text{lead}}^t}) \gets \operatorname{Mean}({{\boldsymbol{\theta}}_{\text{lead}}^t})$\;
    $\nu({{\boldsymbol{\theta}}_{\text{lead}}^t}) \gets \operatorname{Std}({{\boldsymbol{\theta}}_{\text{lead}}^t})$\;
    $\theta _{{\rm{th}}}^t = \max \left( \min \left( {\mu ({\bf{{\boldsymbol{\theta}} }}_{{\rm{lead}}}^t) - \eta \nu ({\bf{{\boldsymbol{\theta}}}}_{{\rm{lead}}}^t),\frac{\pi }{2}} \right), 0 \right)$ ~~~// \textit{Adaptive Threshold}\;
    $\mathcal{{\widehat S}}^t = \left\{ i \in \mathcal{{ S}} \;\middle|\; \theta^{t}_{\text{lead},i} \leq \theta^{t}_{\text{th}} \right\}$\;
}

\Fn{\Fr}{
    \ForEach{$i \in \widehat{\mathcal{S}}^t$}{
        ${\widetilde{\mathcal{L}}}_i^t = \mathcal{L}_i^t + \lambda (1 - \cos\theta_{{\text{lead}}, i}^t)$ ~~~// \textit{Regularized Local Loss}\;
    }
     $    \mathcal{L}^t_{\text{g}} = \sum_{i \in \mathcal{{\widehat S}}^t} {\widetilde{\mathcal{L}}}_i^t$\;
}
\Return{$\mathcal{L}^t_{\text{g}}$}
\end{algorithm}

\subsubsection{Gradient Direction Regularization}\label{sssec:GDA_3}

Building upon the filtering of the adaptive thresholding, we introduce a direction-aware regularization term into the local loss function to rectify the updates of selected clients. We augment the local loss with a nonlinear geometric regularization term $(1 - \cos\theta_{{\text{lead}}, i}^t)$. The adoption of this cosine-based form is motivated by two key properties. First, it maps angular deviation to a smooth, differentiable metric value within $[0, 1]$, ensuring numerical stability during BP. Second, the regularization term exhibits a vanishing characteristic near alignment: as $\theta_{{\text{lead}}, i}^t \to 0$, it naturally decays to zero. This guarantees that once the device gradient is well-aligned, the regularization term vanishes, allowing the model updates to be driven solely by the local loss function. For the $i$-th selected edge device at the $t$-th training round (i.e., $i \in \mathcal{{\widehat S}}^t$, where $\mathcal{{\widehat S}}^t = \left\{ i \in \mathcal{{ S}} \;\middle|\; \theta^{t}_{\text{lead},i} \leq \theta^{t}_{\text{th}} \right\}$), the regularized local loss
is given by
\begin{equation}\label{regu_local_loss}
    {\widetilde{\mathcal{L}}}_i^t = \mathcal{L}_i^t + \lambda (1 - \cos\theta_{{\text{lead}}, i}^t),
\end{equation}
where $\mathcal{L}_i$ denotes the original local loss, $\theta_{{\text{lead}}, i}^t$ is the angular deviation between the $i$-th edge device's gradient and the leader gradient, and $\lambda>0$ represents a regularization coefficient.

{\bf{Remark.}} The design of angular regularization in~\eqref{regu_local_loss} is both adaptive and robust. Since the selected $\theta^{t}_{\text{lead},i}$ is restricted to $[0, \frac{\pi}{2}]$ via the adaptive thresholding in~\eqref{eqn:threshold}, the regularization function exhibits a strict monotonic increase across this domain. As $\theta^{t}_{\text{lead},i} \rightarrow 0$, indicating strong directional alignment, the penalty diminishes to zero and thereby preserves the original update. In contrast, as $\theta^{t}_{\text{lead},i} \rightarrow \frac{\pi}{2}$, the penalty reaches its maximum, effectively suppressing updates that are nearly orthogonal to the global direction.

The global loss function at the $t$-th training round is then formulated by aggregating the regularized local losses over the selected edge devices: 
\begin{equation}
    \mathcal{L}^t_{\text{g}} = \sum_{i \in \mathcal{{\widehat S}}^t} {\widetilde{\mathcal{L}}}_i^t.
\end{equation}

The workflow of GDA is elaborated in {\bf{Algorithm~\ref{alg:GDA}}}.

{\bf{Complexity Analysis.}} Let $d$ be the dimension of each server-side gradient. The LGI module requires $\mathcal{O}(\vert{}\mathcal{S}\vert{}^2 d)$ time to compute pairwise gradient consistency scores and $\mathcal{O}(\vert{}\mathcal{S}\vert{} \log \vert{}\mathcal{S}\vert{})$ time for client ranking. The GDA module takes $\mathcal{O}(\vert{}\mathcal{S}\vert{} d)$ time to compare device gradients against the leader gradient. Therefore, the server-side time complexity is $\mathcal{O}(\vert{}\mathcal{S}\vert{}^2 d)$, dominated by LGI. Notably, these computations consist entirely of lightweight vector dot products rather than intensive matrix multiplications. Given that the number of participating clients in PSL deployments is not very large, this overhead is computationally manageable compared to the standard model backpropagation. By accumulating consistency scores without storing the full pairwise similarity matrix, \name requires $\mathcal{O}(\vert{}\mathcal{S}\vert{} d)$ space to buffer the device gradients. Moreover, LGI and GDA reuse the server-side gradients already available in standard PSL and therefore introduce no additional client–server communication or client-side computation.

\section{{Implementation}}\label{sec:implementation}
In this section, we first elaborate on the implementation of \name and then present the experiment setup.

\begin{figure}[t]
\centering
\includegraphics[width=0.65\linewidth]{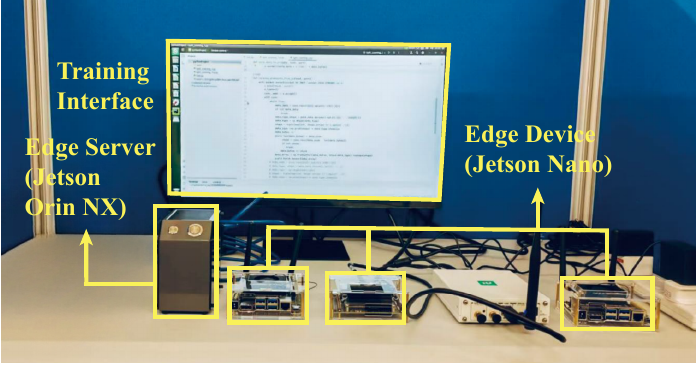}
\caption{\name prototype and testbed.}
\label{fig:testbed_H}
\vspace{-1em}
\end{figure}

\subsection{Implementing \name}

We implement the \name prototype using a micro services architecture to support modular deployment and flexible orchestration of both server-side and client-side components, as shown in Fig.~\ref{fig:testbed_H}. The testbed is constructed using NVIDIA Jetson development kits~\cite{nvidia_jetson2024}. Specifically, the server is implemented on an NVIDIA Jetson Orin NX, featuring a 1024-core Ampere GPU with 32 Tensor Cores and an 8-core ARM Cortex-A78AE v8.2 64-bit CPU, providing support for server-side FP and BP, as well as coordination mechanisms. To emulate edge devices, we adopt NVIDIA Jetson Nano, each equipped with a 128-core Maxwell GPU and a quad-core ARM Cortex-A57 CPU running at 1.43 GHz. Regarding the software stack, we utilize HTTP/2 for the underlying transport protocol to ensure efficient and reliable data transmission and Protocol Buffers for data serialization. The entire system is implemented using Python 3.7, with deep learning tasks executed via PyTorch 1.9.1.

\subsubsection{Dataset and Models}\label{sec:simulation_setup_dataset}

The evaluated datasets and models are selected to fit the memory and computing capacities of our SL testbed (comprising Jetson Orin NX and Jetson Nano), thereby enabling a representative performance evaluation in edge computing systems.
For the datasets, we adopt two widely used image classification datasets: CIFAR-10 and CIFAR-100~\cite{krizhevsky2009learning}.
In our experiment, we consider both IID and non-IID data settings: i) In the IID setting, the global dataset is randomly shuffled and evenly distributed among all participating edge devices, ensuring uniform class distribution and balanced sample sizes; ii) In the non-IID setting, we adopt the Dirichlet data partitioning~\cite{hu2025faster,lin2025leo,zhang2024multi} to set varying degrees of class imbalance controlled by parameter $\alpha$, where a larger $\alpha$ indicates a more balanced class across edge devices. For the network architectures, we evaluate \name using VGG-16~\cite{simonyan2014very} and Vision Transformer-Base (ViT-Base)~\cite{dosovitskiy2020image}.
For model partitioning, we split both models into client-side and server-side models for deployments. For VGG-16, the first four convolutional layers are designated as the client-side model, and the remaining layers are deployed as the server-side model. For ViT-Base, the first three transformer blocks constitute the client-side model, with the subsequent blocks processed by the server as the server-side model.



\subsubsection{Benchmarks}\label{sec:base_line}
To comprehensively evaluate the performance of \name, we compare \name against the following four alternatives: 
\begin{itemize}
\item \textbf{Vanilla SL} is the original version of SL~\cite{vepakomma2018split}, where client-side training is executed sequentially across edge devices. In each training round, a single device collaborates with the server to update the model, and the updated client-side model is then passed to the next device for subsequent training.
\item \textbf{SFL} is the most widely adopted variant of SL that enables parallel client-side training across edge devices, where each device updates its client-side model and uploads it to the parameter server for model aggregation~\cite{thapa2022splitfed}.
\item \textbf{PSL} is a variant of SFL (without client-side model synchronization compared with SFL)~\cite{kim2022bargaining,joshi2021splitfed} that trains a shared server-side model using activations from all edge devices, while each device independently updates its own client-side model based on local datasets.
\item \textbf{EPSL} is a fixed-cut-layer variant of the framework in~\cite{lin2023efficient}, where a portion of the last-layer activation's gradients is aggregated on the server to reduce computational and communication overhead without adaptively adjusting the cut layer.

\end{itemize}

\begin{figure}[t]
  \centering
  \subfloat[{CIFAR-10 on VGG-16 under IID
 setting.} \label{fig:vgg_cifar_iid}]
  {
    \includegraphics[width=4.3cm, height=2.45cm]{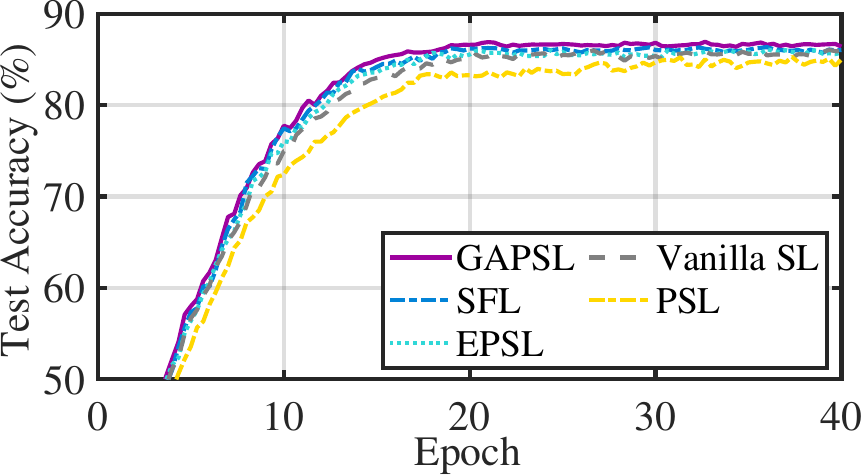}
  }
    \subfloat[{CIFAR-10 on VGG-16 under non-IID
 setting.}\label{fig:vgg_cifar_non_iid} ]
  {
    \includegraphics[width=4.3cm, height=2.44cm]{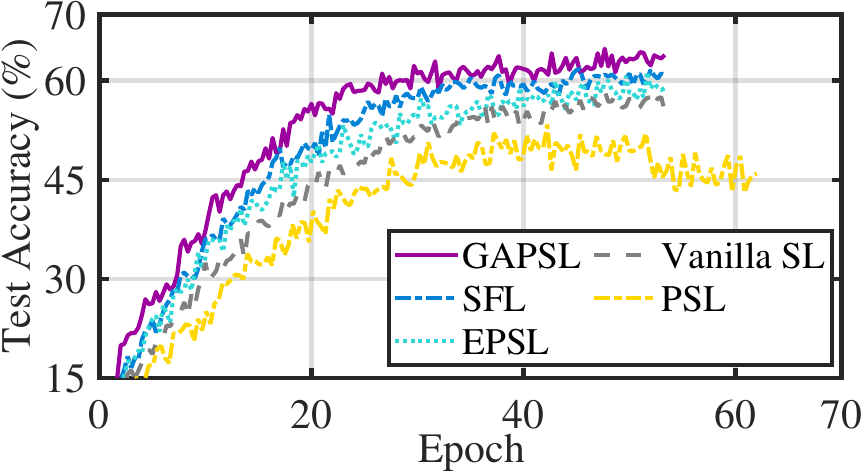}
  } \\
  \subfloat[{CIFAR-100 on ViT-Base under IID
 setting.} \label{fig:vit_cifar_iid}
  ]
  {
    \includegraphics[width=4.39cm, height=2.49cm]{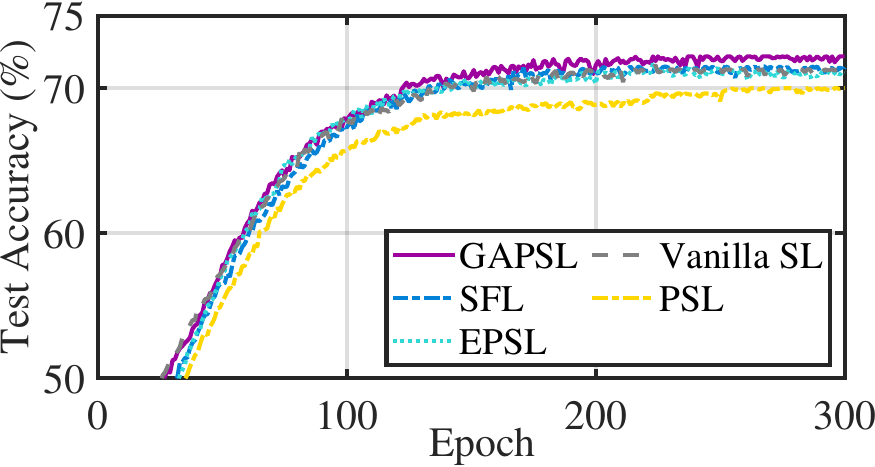}
  }
  \subfloat[{CIFAR-100 on ViT-Base under non-IID
 setting.} \label{fig:vit_cifar_non_iid} ]
  {
    \includegraphics[width=4.3cm, height=2.45cm]{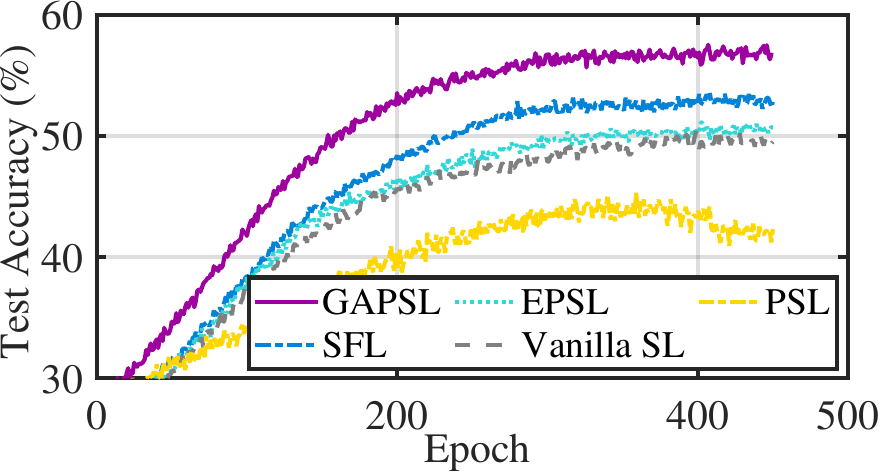}
  }
  \caption{{The training performance for CIFAR-10 and CIFAR-100 under IID and non-IID settings using VGG-16 and ViT-Base.}}
  \label{fig:simulation_overall_curve}
\end{figure}

\subsubsection{Hyper-parameters}\label{sec:hyper-para}

In our experiments, we deploy $S=10$ edge devices, and $\alpha$ is set to 0.1 for the non-IID setting by default unless specified otherwise. The regularization coefficient $\lambda$ is set to $5 \times 10^{-4}$ for VGG-16 and $3 \times 10^{-5}$ for ViT-Base. The gradient selection ratio is is bounded between $K_{\min} = 20\%$ and $K_{\max} = 80\%$. For model-specific configurations, we use a batch size of 128 for VGG-16 and 64 for ViT-Base. The regularization coefficient $\lambda$ is set to $5 \times 10^{-4}$ for VGG-16 and $3 \times 10^{-5}$ for ViT-Base. For VGG-16, optimization is performed using stochastic gradient descent SGD with a momentum of 0.9. The learning rates are set to 0.005 for client-side training and 0.01 for the server. For ViT-Base, we adopt the AdamW optimizer, with learning rates of 0.0001 for clients and 0.0003 for the server. For a fair comparison, all baseline methods strictly share identical hyper-parameters, including the learning rate, batch size, and network architecture.

\section{{Performance Evaluation}}\label{sec:simulation results}

In this section, we evaluate the performance of \name from three aspects: i) comparisons with four benchmarks to demonstrate the superiority of \name; ii) investigating the impact of SL-related hyperparameters on the model performance; iii) ablation studies to show the necessity of each meticulously designed component in \name, including LGI and GDA.

\begin{figure}[t!]
  \centering
  \subfloat[{CIFAR-10 on VGG-16.} \label{fig:vgg_converged_accuracy}]
  {
    \includegraphics[width=4.2cm, height=2.2cm]{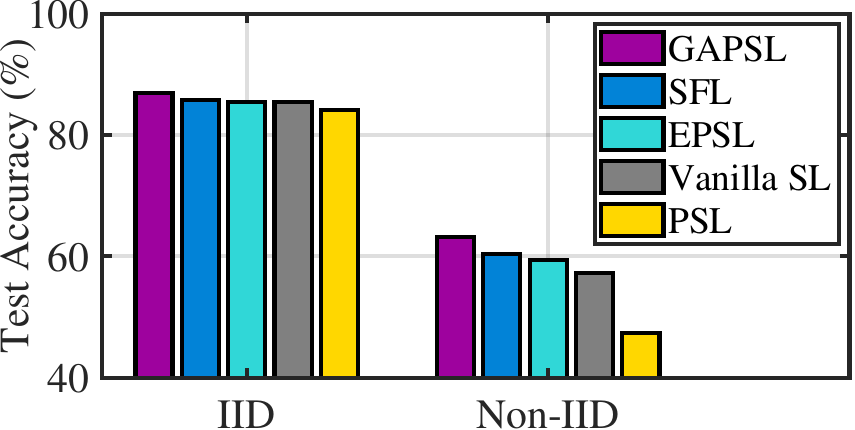}
  }
    \subfloat[{CIFAR-100 on ViT-Base.}\label{fig:vit_converged_accuracy} ]
  {
    \includegraphics[width=4.2cm, height=2.2cm]{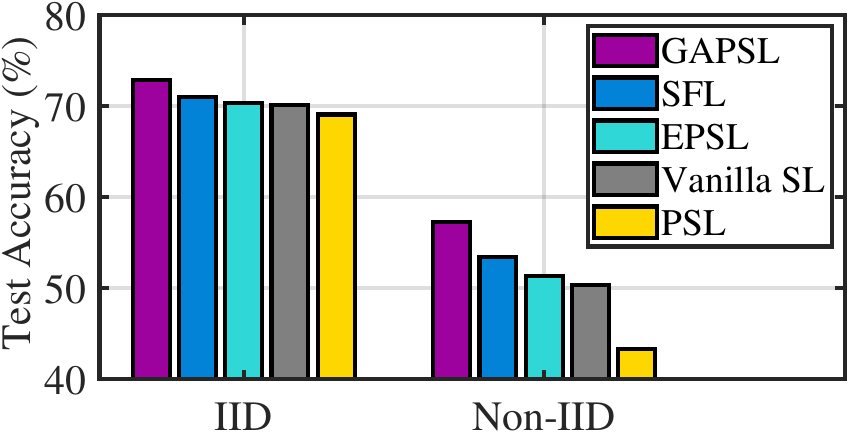}
  } 
  \caption{The converged accuracy for CIFAR-10 and CIFAR-100 under IID and non-IID settings using VGG-16 and ViT-Base.}
  \label{fig:overall_converged_accuracy}
\end{figure}

\begin{figure}[t]
\vspace{-1em}
  \centering
  \subfloat[{CIFAR-10 on VGG-16.} \label{fig:vgg_converged_time}
  ]
  {
    \includegraphics[width=4.3cm, height=2.35cm]{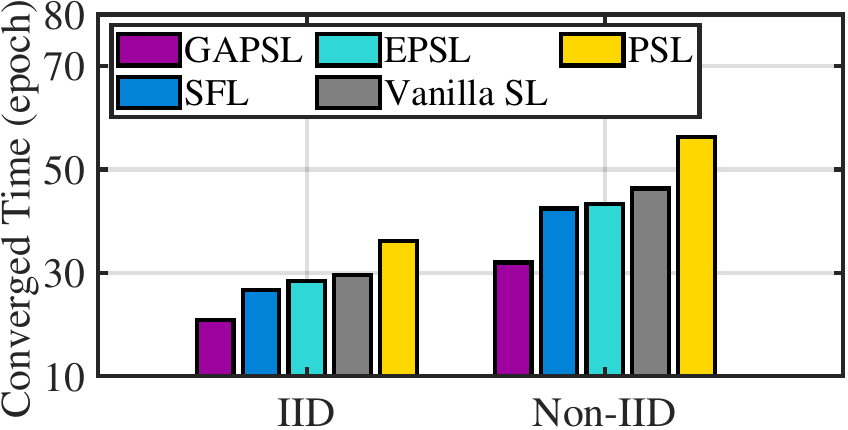}
  }
  \subfloat[{CIFAR-100 on ViT-Base.} \label{fig:vit_converged_time} ]
  {
    \includegraphics[width=4.3cm, height=2.35cm]{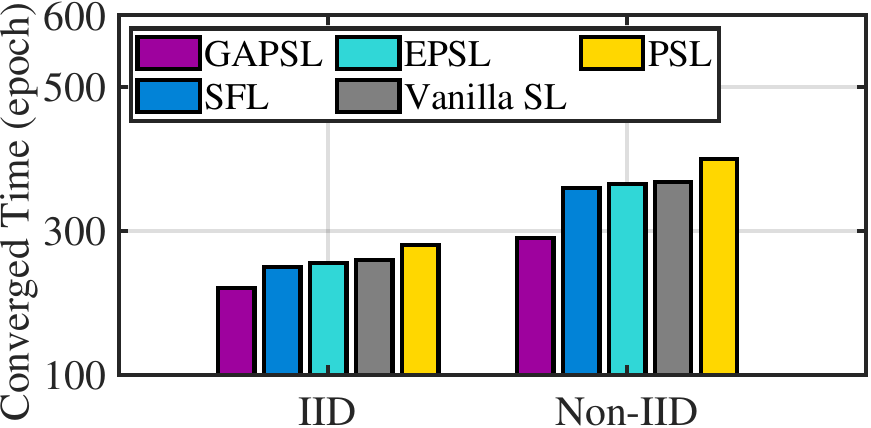}
  }
  \caption{The converged time for CIFAR-10 and CIFAR-100 under IID and non-IID settings using VGG-16 and ViT-Base.}
  \label{fig:overall_converged_time}
\end{figure}

\begin{table}[t!]
\centering
\caption{Test accuracy (\%) and convergence time (epochs) on CIFAR-10 and CIFAR-100.}
\label{tab:fig14_reformatted}
\resizebox{1\linewidth}{!}{%
\begin{tabular}{cccccccc}
\toprule
\textbf{Dataset} & \textbf{Model} & \textbf{Metric (Setting)} & \textbf{GAPSL} & \textbf{SFL} & \textbf{EPSL} & \textbf{Vanilla SL} & \textbf{PSL} \\
\midrule

\multirow{4}{*}{CIFAR-10} & \multirow{4}{*}{VGG-16} 
& Accuracy (IID)        & \cellcolor{gray!20}\textbf{86.9} & 86.2 & 85.8 & 85.7 & 84.3 \\
&                       & Accuracy (Non-IID)    & \cellcolor{gray!20}\textbf{63.3} & 60.7   & 59.4   & 57.3   & 47.3   \\
&                       & Time (IID)            & \cellcolor{gray!20}\textbf{21}   & 27   & 28   & 29   & 36   \\
&                       & Time (Non-IID)        & \cellcolor{gray!20}\textbf{32}   & 42   & 43   & 46   & 56   \\

\midrule

\multirow{4}{*}{CIFAR-100} & \multirow{4}{*}{ViT-Base} 
& Accuracy (IID)        & \cellcolor{gray!20}\textbf{72.1} & 71.5   & 70.8   & 70.7  & 69.8   \\
&                        & Accuracy (Non-IID)   & \cellcolor{gray!20}\textbf{57.0} & 53.4   & 51.4   & 50.3   & 43.2   \\
&                        & Time (IID)           & \cellcolor{gray!20}\textbf{220}  & 254  & 254  & 261  & 283  \\
&                        & Time (Non-IID)       & \cellcolor{gray!20}\textbf{290}  & 361  & 366  & 369  & 401  \\

\bottomrule
\end{tabular}%
}
\vspace{-1.5em}
\end{table}

\subsection{Superiority of \name}

In this section, we conduct a comprehensive comparison of \name against four benchmarks in terms of test accuracy and convergence speed.

\subsubsection{Training Performance of \name} 

Fig.~\ref{fig:simulation_overall_curve} demonstrates the superior performance of \name over four benchmarks on CIFAR-10 and CIFAR-100 datasets under IID and non-IID settings. \name consistently achieves superior performance across both IID and non-IID settings. This performance gain is primarily attributed to the synergistic design of LGI and GDA, which effectively identifies the global optimization direction and rectifies local gradient updates via gradient direction alignment. 
Though vanilla SL is theoretically equivalent to centralized mini-batch SGD, GAPSL slightly outperforms it under IID settings. This improvement stems from the regularization provided by the GDA, which acts as a smoothing filter over the stochastic noise in local mini-batch sampling to mitigate gradient jitter.  
While data heterogeneity induces severe gradient conflicts, \name demonstrates remarkable resilience, effectively resolving gradient inconsistency to maintain robust convergence toward the global optimum. In contrast, vanilla SL and PSL exhibit significant performance degradation, particularly under non-IID settings, due to their lack of statistical gradient calibration. Although SFL and EPSL offer moderate gains through model aggregation or partial synchronization, they still fall short of the efficacy achieved by \name. By comparing Fig.~\ref{fig:vgg_cifar_iid} with Fig.~\ref{fig:vgg_cifar_non_iid}, and Fig.~\ref{fig:vit_cifar_iid} with Fig.~\ref{fig:vit_cifar_non_iid}, we show that \name and the other four benchmarks experience slower convergence under the non-IID setting than the IID setting.

\begin{figure}[t]
  \centering
  \subfloat[{CIFAR-10 on VGG-16 under IID
 setting.} \label{fig:accuracy_num_device_iid}]
  {
    \includegraphics[width=4.3cm, height=2.5cm]{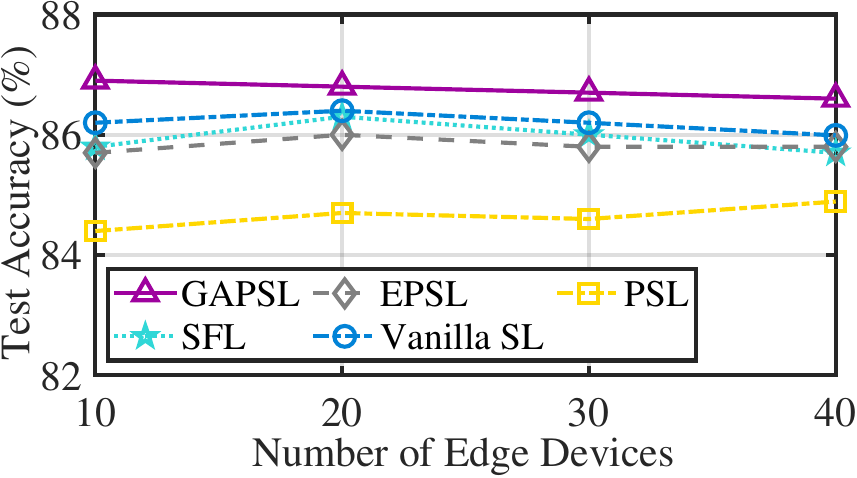}
  }
    \subfloat[{CIFAR-10 on VGG-16 under non-IID
 setting.}\label{fig:accuracy_num_device_non_iid} ]
  {
    \includegraphics[width=4.3cm, height=2.5cm]{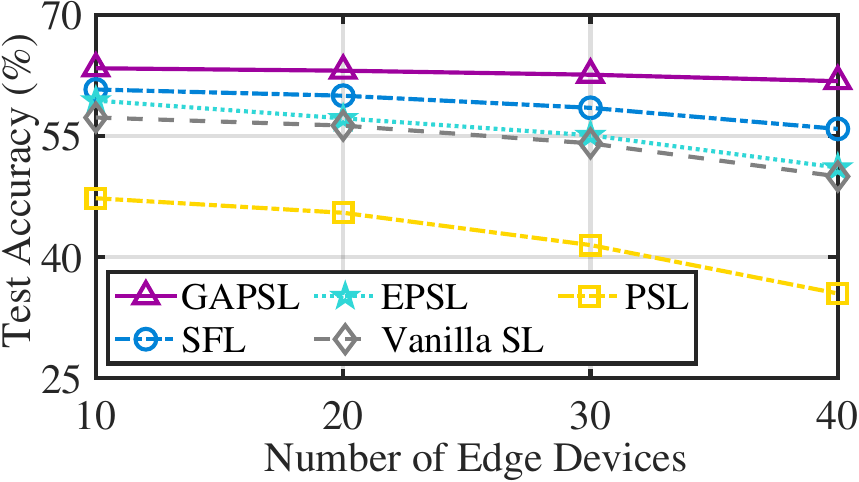}
  } 
  \caption{The test accuracy versus number of edge devices on CIFAR-10 under IID and non-IID settings using VGG-16.}
  \label{fig:overall_accuracy_num_device}
\end{figure}

\subsubsection{Converged Accuracy of \name}

Fig.~\ref{fig:overall_converged_accuracy} illustrates the converged accuracy of \name against four other benchmarks on CIFAR-10 and CIFAR-100 datasets. In the non-IID setting, both PSL and vanilla SL suffer from substantial accuracy degradation, with an accuracy drop of approximately 37.0\% (resp. 26.6\%) and 28.4\% (resp. 20.4\%) on CIFAR-10 (resp. CIFAR-100) relative to their IID counterparts. This degradation in PSL is attributed to the lack of gradient coordination, which induces directional inconsistency and suboptimal updates. Vanilla SL is plagued by catastrophic forgetting inherent to its sequential training paradigm, where round-robin updates over heterogeneous data lead to overfitting on recent samples and the erasure of prior knowledge. While SFL and EPSL mitigate these issues via partial aggregation or synchronization, they still incur noticeable accuracy losses due to their limited capability in resolving severe gradient conflicts. In contrast, \name demonstrates superior robustness, outperforming SFL, EPSL, vanilla SL, and PSL by margins of 2.6\% (resp. 3.6\%), 3.9\% (resp. 5.6\%), 6.0\% (resp. 6.7\%), and 16.0\% (resp. 13.8\%) on CIFAR-10 (resp. CIFAR-100) under the non-IID setting. This significant improvement validates the effectiveness of our design, where LGI extracts directionally consistent trends, and GDA enforces rigorous alignment across edge devices.


\subsubsection{Converged Time of \name}

Fig.~\ref{fig:overall_converged_time} compares the converged time of \name against four other benchmarks on CIFAR-10 and CIFAR-100 datasets. In the IID setting, SFL, EPSL, vanilla SL, and PSL take approximately 1.3× (resp. 1.2×), 1.3× (resp. 1.2×), 1.4× (resp. 1.2×), and 1.7× (resp. 1.3×) more epochs to converge than \name on CIFAR-10 (resp. CIFAR-100). This notable performance gain is attributed to the synergistic design of LGI and GDA. Specifically, LGI effectively filters out divergent updates to extract a lead gradient that reflects the consensus reference direction, while GDA rigorously aligns local gradients with this reference. This selective and adaptive design effectively suppresses gradient noise and expedites model convergence. In contrast, vanilla SL exhibits the slowest convergence due to its inherently sequential training paradigm. SFL, EPSL, and PSL suffer from prolonged training latencies caused by inconsistent gradient directions across devices. In the non-IID setting, where data heterogeneity exacerbates gradient divergence and optimization drift, all methods experience increased convergence times. Specifically, PSL requires more communication rounds to converge, offsetting its advantage of a shorter per-round delay. \name explicitly overcomes this limitation: by aligning gradient directions via LGI and GDA, it effectively accelerates convergence and reduces the required communication rounds. The experimental results demonstrate superior robustness and adaptability to statistical heterogeneity. Detailed numerical results are summarized in Table~\ref{tab:fig14_reformatted}.

\begin{figure}[t!]
  \centering
  \subfloat[{CIFAR-10 on VGG-16.} \label{fig:vgg_data_hetero}]
  {
    \includegraphics[width=4.3cm, height=2.45cm]{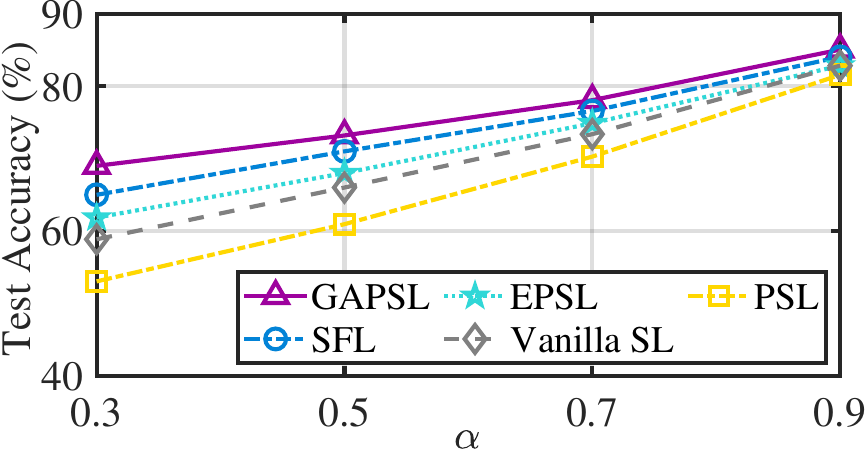}
  }
    \subfloat[{CIFAR-100 on ViT-Base.}\label{fig:vit_data_hetero} ]
  {
    \includegraphics[width=4.3cm, height=2.45cm]{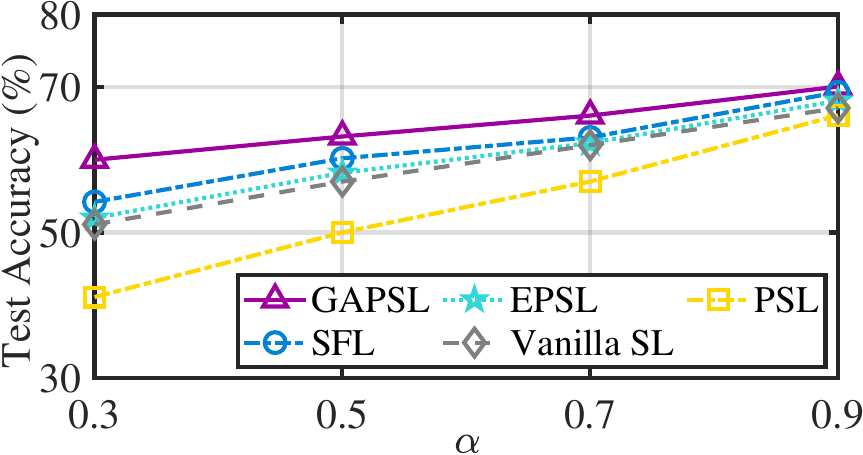}
  } 
  \caption{The test accuracy versus degrees of data  heterogeneity on CIFAR-10 and CIFAR-100 using VGG-16 and ViT-Base.}
  \label{fig:overall_data_hetero}
\end{figure}

\begin{table}[t]
\centering
\caption{Test accuracy (\%) versus number of devices on CIFAR-10 using VGG-16.}
\label{tab:fig15_reformatted}
\resizebox{0.93\linewidth}{!}{%
\begin{tabular}{ccccccc}
\toprule
\textbf{Setting} & \textbf{Devices Scale} & \textbf{GAPSL} & \textbf{SFL} & \textbf{EPSL} & \textbf{Vanilla SL} & \textbf{PSL} \\
\midrule

\multirow{4}{*}{IID}
& 10 & \cellcolor{gray!20}\textbf{86.9} & 86.2 & 85.8 & 85.7 & 84.3 \\
& 20 & \cellcolor{gray!20}\textbf{86.8} & 86.4 & 86.3 & 86.1 & 84.7 \\
& 30 & \cellcolor{gray!20}\textbf{86.7} & 86.2 & 86.0 & 85.8 & 84.6 \\
& 40 & \cellcolor{gray!20}\textbf{86.5} & 85.9 & 85.7 & 85.7 & 84.9 \\

\midrule

\multirow{4}{*}{Non-IID}
& 10 & \cellcolor{gray!20}\textbf{63.3} & 60.7   & 59.4   & 57.3   & 47.3 \\
& 20 & \cellcolor{gray!20}\textbf{63.1} & 60.1 & 57.2 & 56.3 & 45.5 \\
& 30 & \cellcolor{gray!20}\textbf{62.6} & 58.5 & 55.1 & 54.2 & 41.4 \\
& 40 & \cellcolor{gray!20}\textbf{61.7} & 55.9 & 55.3 & 50.1 & 35.6 \\

\bottomrule
\end{tabular}%
}
\end{table}

\subsection{Micro-benchmarking}

In this section, we investigate the impact of hyperparameters on the model performance.

\subsubsection{Impact of Number of Edge Devices}

Fig.~\ref{fig:overall_accuracy_num_device} shows the converged accuracy of \name and other benchmarks versus the number of edge devices on CIFAR-10 and CIFAR-100 datasets under IID and non-IID settings. \name consistently achieves the highest test accuracy across varying numbers of edge devices, demonstrating strong scalability and robustness to the device scale. In the IID setting, all benchmarks exhibit marginal accuracy changes as the number of devices increases, benefiting from the uniform data distribution across edge devices. Nevertheless, \name illustrates the highest accuracy, attributed to its direction-aware gradient alignment mechanism that preserves cross-device gradient consistency. In contrast, all benchmarks suffer from significant accuracy drops with increasing device scale in the non-IID setting, which is caused by the exacerbated gradient inconsistency on highly heterogeneous data. However, \name exhibits remarkable resilience, showing only a negligible accuracy decline even as the system scales to 40 edge devices. This robustness is underpinned by the synergistic design of LGI and GDA that effectively mitigates the gradient inconsistency caused by data heterogeneity. These observations validate the superior scalability of \name for large-scale edge computing systems. Detailed numerical results are summarized in
Table~\ref{tab:fig15_reformatted}.


\subsubsection{Impact of Data Heterogeneity}

Fig.~\ref{fig:overall_data_hetero} illustrates the converged accuracy of \name and other benchmarks versus the degree of data heterogeneity on CIFAR-10 and CIFAR-100 datasets. It can be seen that all benchmarks experience a noticeable decline in model accuracy as the degree of data heterogeneity increases (indicated by a decrease in $\alpha$). Specifically, PSL exhibits the most severe performance degradation, with accuracy dropping by up to 27.8\% (resp. 24.9\%) on CIFAR-10 (resp. CIFAR-100) as $\alpha$ decreases from 0.9 to 0.3. This sharp decline stems from its lack of client-side aggregation, which exacerbates gradient divergence in the non-IID setting. While SFL and EPSL exhibit moderate accuracy loss, they benefit partially from model aggregation or partial gradient synchronization. In contrast, \name consistently maintains superior accuracy across varying data heterogeneity. Even in a highly non-IID setting, \name incurs only a minimal accuracy reduction. This superior resilience showcases the effectiveness of \name’s design in mitigating the adverse effects of data heterogeneity, underscoring its scalability and stability on heterogeneous data. Detailed numerical results are summarized in
Table~\ref{tab:fig16_reformatted}.

\begin{figure}[t]
  \centering
  \subfloat[{CIFAR-10 on VGG-16.} \label{fig:vgg_aba_LGI}]
  {
    \includegraphics[width=4.3cm, height=2.45cm]{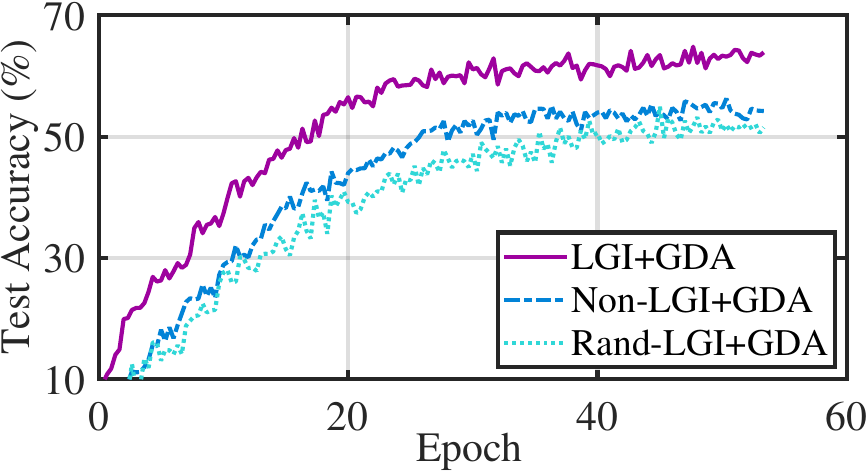}
  }
    \subfloat[{CIFAR-100 on ViT-Base.}\label{fig:vit_aba_LGI} ]
  {
    \includegraphics[width=4.3cm, height=2.45cm]{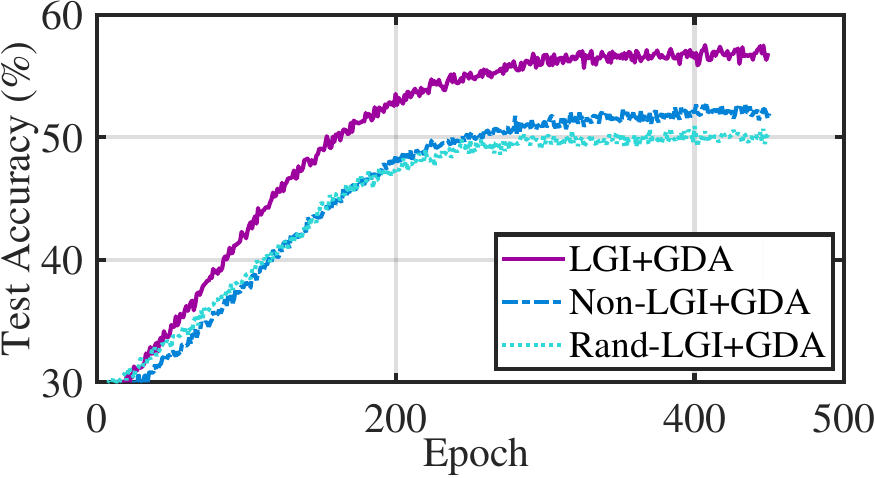}
  } 
  \caption{The ablation evaluation for LGI on CIFAR-10 and CIFAR-100 under non-IID setting using VGG-16 and ViT-Base.}
  \label{fig:overall_aba_LGI}
\end{figure}


\subsection{Ablation Study}

This section evaluates the impact of each module of \name on the model performance.

\subsubsection{Leader Gradient Identification}

Fig.~\ref{fig:overall_aba_LGI} shows the impact of LGI on training performance under the non-IID setting. We compare the training performance of LGI with two ablated variants: Non-LGI, which aggregates gradients from all devices to construct a leader gradient without gradient filtering, and Rand-LGI, which constructs the leader gradient via random gradient subset selection. The results demonstrate that LGI consistently yields superior test accuracy compared to both variants. This advantage stems from LGI’s directional consistency-aware gradient selection, which dynamically identifies and aggregates only directionally consistent gradients. In contrast, Non-LGI suffers from noticeable performance degradation due to the inclusion of conflicting gradients that destabilize the update direction. Rand-LGI exhibits even worse performance, indicating that random selection fails to capture the underlying convergence trend, leading to unstable updates and poor generalization. Notably, the performance gap between LGI and its ablated variants is more pronounced in the ViT-Base model than in VGG-16. This suggests that deeper and more complicated transformer-based architectures are more sensitive to gradient inconsistency and thus benefit more significantly from the robust gradient filtering provided by LGI.

\begin{figure}[t]
  \centering
  \subfloat[{CIFAR-10 on VGG-16.} \label{fig:vgg_aba_GDA}]
  {
    \includegraphics[width=4.3cm, height=2.45cm]{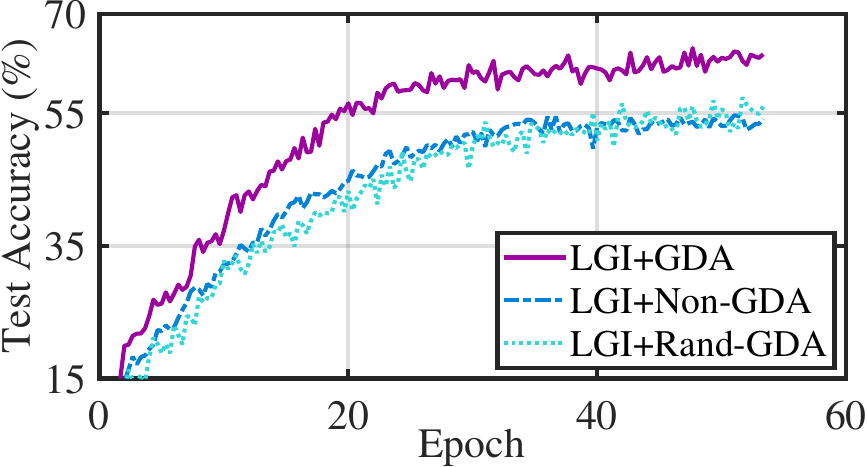}
  }
    \subfloat[{CIFAR-100 on ViT-Base.}\label{fig:vit_aba_GDA} ]
  {
    \includegraphics[width=4.3cm, height=2.45cm]{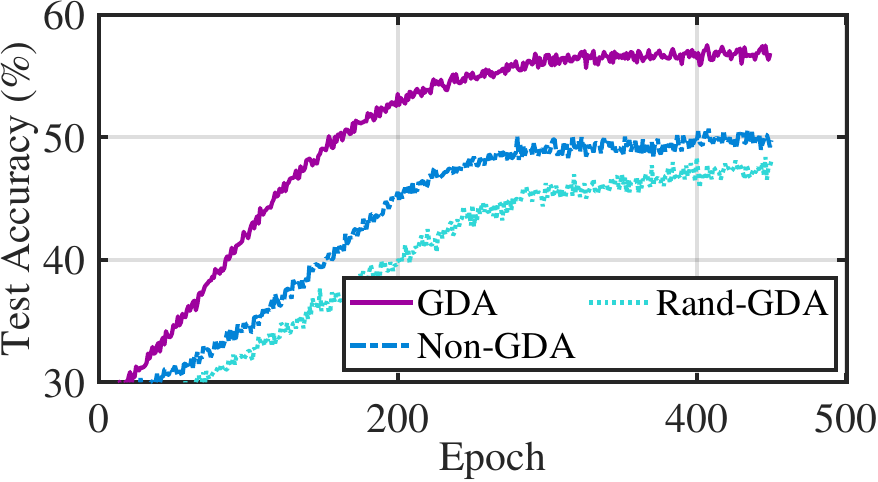}
  } 
  \caption{The ablation evaluation for GDA on CIFAR-10 and CIFAR-100 using VGG-16 and ViT-Base under non-IID setting.}
  \label{fig:overall_aba_GDA}
\end{figure}


\begin{table}[t]
\centering
\caption{Test accuracy (\%) versus data heterogeneity (Dirichlet $\alpha$) on CIFAR-10 and CIFAR-100.}
\label{tab:fig16_reformatted}
\resizebox{0.97\linewidth}{!}{%
\begin{tabular}{lllccccc}
\toprule
\textbf{Dataset} & \textbf{Model} & $\boldsymbol{\alpha}$ & \textbf{GAPSL} & \textbf{SFL} & \textbf{EPSL} & \textbf{Vanilla SL} & \textbf{PSL} \\
\midrule

\multirow{4}{*}{CIFAR-10} & \multirow{4}{*}{VGG-16}
& 0.9 & \cellcolor{gray!20}\textbf{85.0} & 84.1 & 82.8 & 82.7 & 81.3 \\
& & 0.7 & \cellcolor{gray!20}\textbf{78.1} & 76.6 & 74.9 & 73.4 & 70.3\\
& & 0.5 & \cellcolor{gray!20}\textbf{73.2} & 71.2 & 68.0 & 66.2 & 60.9 \\
& & 0.3 & \cellcolor{gray!20}\textbf{69.3} & 65.1 & 61.8 & 58.9 & 53.5 \\

\midrule

\multirow{4}{*}{CIFAR-100} & \multirow{4}{*}{ViT-Base}
& 0.9 & \cellcolor{gray!20}\textbf{70.1} & 69.3 & 68.2 & 67.3 & 66.1 \\
& & 0.7 & \cellcolor{gray!20}\textbf{66.2} & 63.1 & 62.3 & 62.0 & 57.4 \\
& & 0.5 & \cellcolor{gray!20}\textbf{63.4} & 60.2 & 58.2 & 57.1 & 50.0 \\
& & 0.3 & \cellcolor{gray!20}\textbf{60.6} & 54.3 & 52.1 & 51.4 & 41.2 \\

\bottomrule
\end{tabular}%
}
\vspace{-1.5em}
\end{table}

\subsubsection{Gradient Direction Alignment}

Fig.~\ref{fig:overall_aba_GDA} shows the impact of GDA on training performance under the non-IID setting. We compare GDA with two ablated variants: Non-GDA, which removes the gradient alignment mechanism and relies solely on the leader gradient for model update, and Rand-GDA, which employs gradient alignment randomly to a subset of local gradients, regardless of their angular deviation from the leader gradient. The results demonstrate that GDA exhibits the highest test accuracy, which is attributed to GDA’s adaptive regularization, which imposes a geometric constraint on each device’s loss function to align its gradients with the consensus reference direction. In contrast, Non-GDA exhibits the lowest accuracy due to the absence of alignment, which causes local gradients to deviate significantly from the global objective. Rand-GDA performs marginally better, but its random alignment fails to ensure consistent updates, resulting in unstable optimization and reduced generalization. These observations confirm the essential role of GDA in improving convergence stability and enhancing model performance.

\section{Related Work}\label{sec:related_work}

{\textit{1) Split Learning.}} SL has emerged as a promising privacy-enhancing distributed learning paradigm to address the limitations of FL. Vepakomma~\textit{et al.}~\cite{vepakomma2018split} first proposed the original SL, known as vanilla SL, where clients collaborate with the server in a round-robin manner for co-training; however, this client-side sequential training manner suffers from excessive training latency. To enable client-side parallel training, Thepa~\textit{et al.}~\cite{thapa2022splitfed} developed the most prevalent SL variant, coined SFL, that integrates model partitioning of SL with parallel training of FL to parallelize the client-side model training, enabling multiple devices to collaboratively train a shared model. Building upon this, Kim~\textit{et al.} and Joshi~\textit{et al.}~\cite{kim2022bargaining,joshi2021splitfed} devised the PSL framework that eliminates client-side model aggregation to streamline the training procedure of SFL, accelerating model convergence while compromising the training performance. Subsequent research endeavors have been dedicated to developing more communication- and computation-efficient SL. Lin~\textit{et al.}~\cite{lin2023efficient} designed an efficient PSL framework that employs last-layer gradient aggregation to shrink the dimensionality of activations’ gradients for reducing the training latency. Pal~\textit{et al.}~\cite{pal2021server} devised a more scalable PSL framework that averages the local gradients at the cut layer to overcome server-side large batch and the backward client decoupling problems. Lin~\textit{et al.}~\cite{lin2024adaptsfl,lin2024hierarchical} derived a theoretical convergence analysis for SFL and proposed an efficient algorithm to adaptively control model splitting and sub-model aggregation interval.

{\textit{2) Data Heterogeneity Mitigation.}} Research efforts to mitigate data heterogeneity in distributed learning predominantly focus on FL, with two major research directions: {model aggregation optimization}, which aims to reduce update bias in model aggregation, and {data-adaptive personalization}, which seeks to tailor models to individual devices’ data distributions instead of forcing a single global model to fit all. For \textit{model aggregation optimization}, Shi~\textit{et al.}~\cite{shi2025fedawa} proposed FedAWA that adaptively adjusts aggregation weights based on the direction of client model updates, eliminating the need for additional data and thereby enhancing both the stability and generalization of the global model. Karimireddy~\textit{et al.}~\cite{karimireddy2020scaffold} developed SCAFFOLD that controls variance reduction in model aggregation to reduce client-drift, thus achieving model convergence with fewer communication rounds. Deng~\textit{et al.}~\cite{deng2024fedasa} proposed a cell-wise shared model layer selection to adaptively construct the shared architecture across devices, followed by a cell-based aggregation strategy tailored to device-specific architectures. For \textit{data-adaptive personalization},  Wu~\textit{et al.}~\cite{wu2024fedcache} devised FedCache that reserves a server-side knowledge cache for fetching personalized knowledge from the samples with similar hashes and then employs ensemble distillation to personalize local models for local bias mitigation. Hu~\textit{et al.}~\cite{hu2025faster} designed a clustered data sharing FL framework that selectively shares partial data from cluster heads to trusted associates via sidelink multicasting to mitigate data heterogeneity. Wang~\textit{et al.}~\cite{wang2023towards} proposed pFedHR, which generates diverse personalized models and applies model reassembly to alleviate the adverse effects of data heterogeneity.

While the above frameworks have demonstrated effectiveness under data heterogeneity, they typically rely on full model aggregation and frequent global synchronization, which is inherently incompatible with the PSL that eliminates client-side aggregation via model partitioning. Recent studies have addressed data heterogeneity in Split Learning (SL). For example, GPSL~\cite{kohankhaki2025parallel} reduces the impact of non-IID data using a global sampling strategy to balance data distributions, while FSL-SAGE~\cite{nair2025fsl} accelerates federated split learning through smashed activation gradient estimation. However, these methods either require specific client-side data sampling rules or depend on client-side model aggregation to correct data biases. In contrast, \name differs by regularizing the local loss in aggregation-free PSL. Instead of changing data distributions or relying on client-side synchronization, GAPSL corrects the update direction by dynamically aligning inter-device gradients during backpropagation. This directional constraint filters out conflicting updates to ensure stable convergence.


\section{Conclusion}\label{sec:conclusion}

Taking an important step towards the SL paradigm, we have proposed a cutting-edge gradient-aligned PSL framework, named \name, enabling efficient model training without client-side model aggregation. 
\name comprises two key components: leader gradient
identification and gradient direction alignment. 
First, the leader gradient identification mechanism selects and aggregates a subset of directionally consistent device gradients to construct a leader gradient reflecting the consensus reference direction. Second, the gradient direction alignment scheme aligns each device’s gradient with the leader gradient via a direction-aware regularization term to improve model convergence. 
The experimental results demonstrate that \name achieves superior performance compared to state-of-the-art benchmarks.

\ifCLASSOPTIONcaptionsoff
  \newpage
\fi

\bibliographystyle{IEEEtran}
\bibliography{reference}

\end{document}